\RequirePackage{fix-cm}
\documentclass[twocolumn]{svjour3}          
\smartqed  
\usepackage{graphicx}
\usepackage{natbib}
\usepackage{multirow}
\usepackage{amsmath}
\usepackage{amssymb}
\usepackage{bbm}
\usepackage{bm}
\usepackage{booktabs}
\usepackage{array} 
\usepackage{blindtext}
\usepackage{comment}

\newcommand{\tableCellHeight}{1.1}
\newcommand{\tabstyle}[1]{
  \setlength{\tabcolsep}{#1}
  \renewcommand{\arraystretch}{\tableCellHeight}
  \centering
  \footnotesize
}

\usepackage{pifont}

\usepackage[dvipsnames]{xcolor}
\usepackage{color, colortbl}
\definecolor{citecolor}{HTML}{0071bc}
\definecolor{tabhighlight}{HTML}{e5e5e5}
\usepackage[colorlinks,citecolor=citecolor,breaklinks=true]{hyperref}

\makeatletter
\renewcommand\paragraph{
  \@startsection{paragraph} 
  {4} 
  {\z@} 
  {.5em \@plus1ex \@minus.2ex} 
  {-.5em} 
  {\normalfont\normalsize\bfseries} 
}
\makeatother
\usepackage{pifont}
\usepackage{nicefrac}
\usepackage{subcaption}
\usepackage{multirow}
\usepackage{booktabs}
\usepackage{color, colortbl}
\usepackage{bbm}

\def\textBF#1{\sbox\CBox{#1}\resizebox{\wd\CBox}{\ht\CBox}{\textbf{#1}}}

\newsavebox\CBox
\def\textBF#1{\sbox\CBox{#1}\resizebox{\wd\CBox}{\ht\CBox}{\textbf{#1}}}

\newcommand{\rotbox}[1]{\rotatebox{90}{#1}}

\begin{document}
\sloppy

\title{Calibrated Cache Model for Few-Shot Vision-Language Model Adaptation
}


\author{Kun Ding         \and
        Qiang Yu \and
        Haojian Zhang \and
        Gaofeng Meng \and
        Shiming Xiang
}


\institute{
	       Kun Ding \at
              MAIS, Institute of Automation, Chinese Academy of Sciences, China \\
              \email{kun.ding@ia.ac.cn}
           \and
           Qiang Yu \at
              Research Center of Aerospace Information, Institute of Automation, Chinese Academy of Sciences, China \\
              \email{qiang.yu@ia.ac.cn}
           \and
           Haojian Zhang\at
           Engineering Laboratory for Intelligent Industrial Vision, Institute of Automation, Chinese Academy of Sciences, China \\
           \email{zhanghaojian2014@ia.ac.cn}
           \and
           Gaofeng Meng \at
           MAIS, Institute of Automation, Chinese Academy of Sciences, China \\
           \email{gfmeng@nlpr.ia.ac.cn}
           \and
           Shiming Xiang \at
           MAIS, Institute of Automation, Chinese Academy of Sciences, China \\
           \email{smxiang@nlpr.ia.ac.cn}
}

\date{Received: date / Accepted: date}

\maketitle

\begin{abstract}
The few-shot adaptation of contrastive vision-language models (VLMs) like Contrastive Language-Image Pre-training (CLIP) to various downstream tasks has garnered significant attention. Currently, cache-based approaches stand out as both effective and efficient for adapting VLMs, seamlessly integrating the knowledge from training images into the logits function without resorting to intricate prompt optimization. Nonetheless, the cache model in these methods overlooks three crucial aspects. \textit{First}, pre-trained VLMs like CLIP are optimized primarily for image-text similarity, neglecting the importance of image-image similarity, thus leading to a gap between pre-training and cache-based adaptation. \textit{Second}, the current cache model is based on the Nadaraya-Watson (N-W) estimator, which disregards the intricate relationships among training samples while constructing weight function. \textit{Third}, under the condition of limited samples, the logits generated by cache model are of high uncertainty, directly using these logits without accounting for the confidence could be problematic. This work presents three calibration modules aimed at addressing the above challenges. \textit{Similarity Calibration} refines the image-image similarity by using unlabeled images. We add a learnable projection layer with residual connection on top of the pre-trained image encoder of CLIP and optimize the parameters by minimizing self-supervised contrastive loss. \textit{Weight Calibration} introduces a precision matrix into the weight function to adequately model the relation between training samples, transforming the existing cache model to a Gaussian Process (GP) regressor, which could be more accurate than N-W estimator. \textit{Confidence Calibration} leverages the predictive variances computed by GP Regression to dynamically re-scale the logits of cache model, ensuring that the cache model's outputs are appropriately adjusted based on their confidence levels. Besides, to reduce the high complexity of GPs, we propose a group-based learning strategy, which only involves computing the inverse covariance matrix within sampled classes of each group. Integrating the above designs, we first propose a training-free GP-based cache model named GPCache, and subsequently extend it to training-required case. By extensive experiments on 11 few-shot classification datasets, we confirm that the proposed methods can achieve state-of-the-art performance.
\end{abstract}

\section{Introduction}\label{sec:intro}
Vision-language models particularly those grounded in contrastive learning have gained significant attention in recent years. Among these, Contrastive Language-Image Pre-training (CLIP) stands as a pioneering approach, leveraging an immense dataset comprising 400 million image-text pairs to learn robust representations for both modalities. The representation space acquired by CLIP has three important attributes: cross-modal alignment, semantic richness and strong generalization ability. Due to these strengths, CLIP has found extensive applications in tasks such as cross-modal retrieval~\citep{VoP}, visual question answering~\citep{ShenLTBRCYK22}, semantic segmentation~\citep{DenseCLIP,ZegCLIP}, zero-shot/few-shot recognition~\citep{CoOp,TipAdapter,APE,CALIP}. To tackle the challenge of zero-shot image classification, CLIP innovatively computes classifier weights by inputting class names into its text encoder. To effectively bridge the gap between pre-trained and downstream text data, a straightforward prompt template, such as ``A photo of a \{CLASS\}", is employed, where ``{CLASS}" is dynamically substituted with various class names. The final decision logits are then derived by assessing the similarities between the image features and these dynamically generated weights, offering a powerful and flexible solution for zero-shot classification.

Going beyond zero-shot learning, recent researches have focused more on adapting CLIP with minimal training samples in a few-shot setting. In this context, parameter-efficient fine-tuning emerges as a pivotal task. One of the prevalent approaches is prompt tuning, which transplants the concept of prefix tuning~\citep{PrefixTuning,SuWQCLWWLLL0SZ22} from Natural Language Processing (NLP) into the vision domain. CoOp~\citep{CoOp}, a pioneering work in this realm, introduces learnable prefix vectors that are inserted into the token sequence of input texts. These vectors are subsequently optimized specifically for downstream datasets, demonstrating remarkable effectiveness. Subsequently, CoCoOp~\citep{cocoop} enhances the generalization ability of CoOp by leveraging image features to dynamically generate instance-conditioned prompts. Despite their notable achievements, prompt tuning methods are computationally intensive, as the gradients should be backpropagated to the input side.

Another line of work avoids learning prompts and introduces cache model to provide prior information~\citep{TipAdapter,TipX,APE}. The cache model comprises key-value pairs, where the keys are pre-computed image features by CLIP's image encoder and the values are binary image labels. For a test image, its extracted feature vector acts as a query, and the cache model computes the similarities between this query and the stored keys to weight the corresponding labels. The resulting logit vector from the cache model offers complementary information to the original logits generated from text features. Tip-Adapter~\citep{TipAdapter}, a pioneering work of cache-based approach, exemplifies the efficiency of this method as it requires no training. To further enhance performance, a variant called Tip-Adapter-F is proposed, which optimizes the cache keys through training. \cite{APE} noticed that not all features matter for downstream applications, as such they proposed a prior refinement module to decouple domain-specific knowledge. Additionally, they explored the trilateral affinities between test image, cache model and textual representation. They proposed a training-free variant termed APE and a training-required variant termed APE-T. \textbf{Although the cache model plays a vital role in these methods, it is still overly simplified, limiting its ability.}

\begin{figure}[!t]
	\centering
	\begin{subfigure}{.475\columnwidth}
		\centering
		\includegraphics[width=1\linewidth]{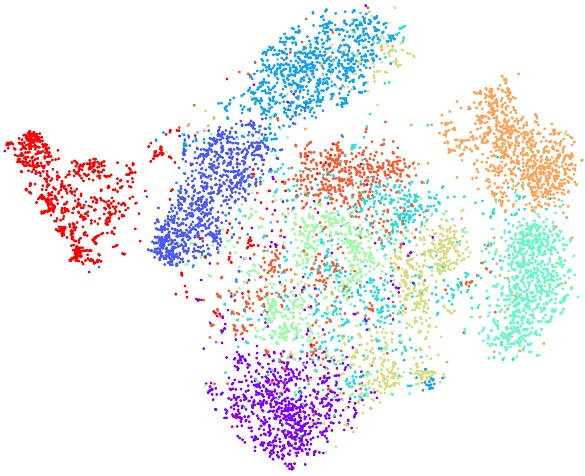}
		\caption{}
		\label{fig:sub1}
	\end{subfigure}%
	\hspace{3pt}
	\begin{subfigure}{.47\columnwidth}
		\centering
		\scalebox{1}[-1]{\includegraphics[width=1\linewidth]{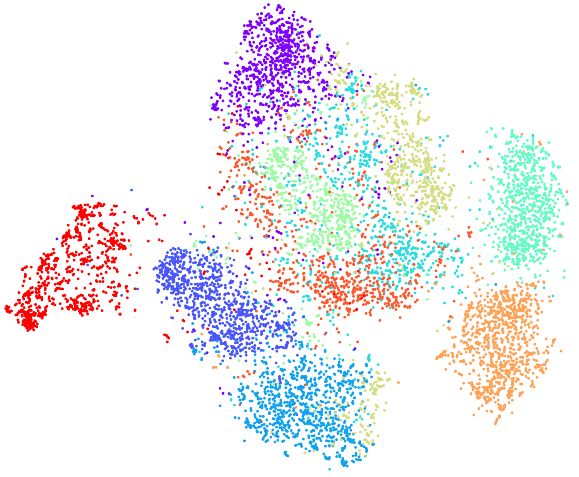}}
		\caption{}
		\label{fig:sub2}
	\end{subfigure}
	\caption{Feature visualization on the test set of EuroSAT dataset without (a) and with (b) similarity calibration.}
	\label{fig:vis_sim_calib}
	\vspace{0pt}
\end{figure}

\begin{figure}[!t]
	\centering
	\begin{subfigure}{.47\columnwidth}
		\centering
		\includegraphics[width=1\linewidth]{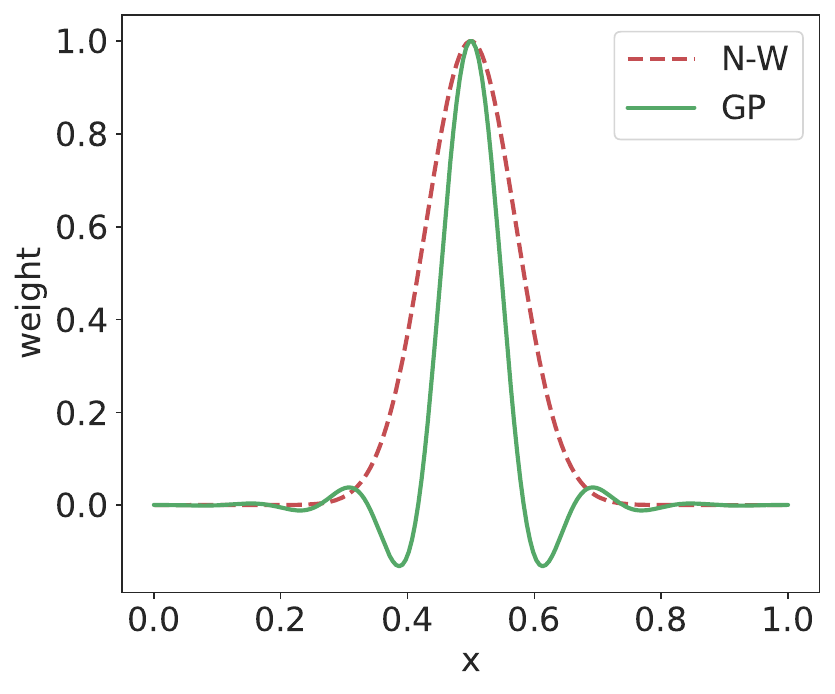}
		\caption{}
		\label{fig:sub11}
	\end{subfigure}
	\hspace{3pt}
	\begin{subfigure}{.47\columnwidth}
		\centering
		\includegraphics[width=1\linewidth]{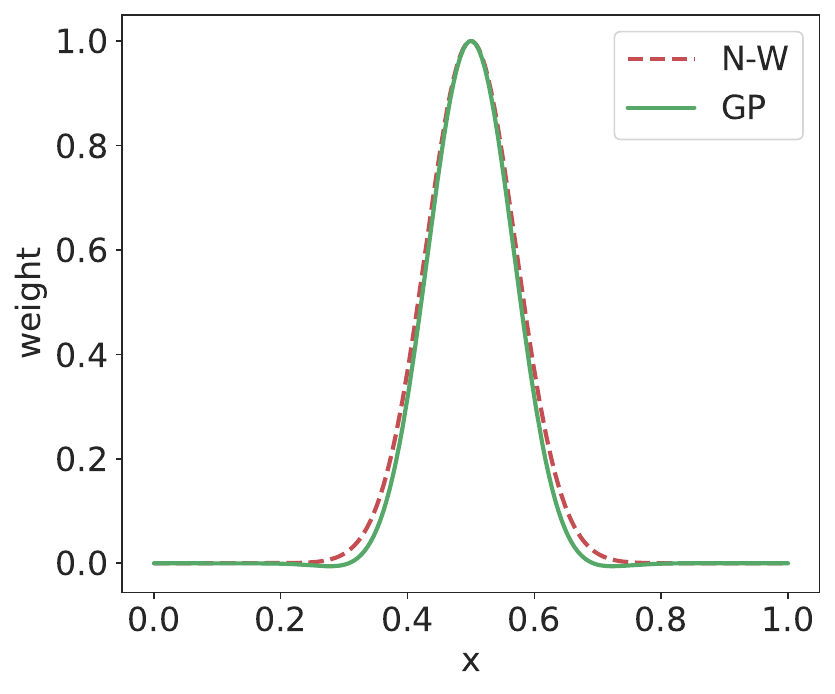}
		\caption{}
		\label{fig:sub21}
	\end{subfigure}
	\caption{Visualization of weight function. The Gaussian kernel $\exp(-\beta x^2)$ with $\beta=100$ is adopted. For GP, 200 uniformly spaced 1D data points between [0,1] are used. The noise variance of GP in (a) and (b) is $\sigma^2=1.0$ and $\sigma^2=100.0$, respectively.}
	\label{fig:weight_function}
	\vspace{-8pt}
\end{figure}

\textbf{First}, existing contrastive learning based VLMs such as CLIP primarily optimize for image-text similarity through cross-modal contrastive loss. This enables them to excel in comparing the differences and similarities between images and texts but inherently limits their proficiency in distinguishing between images. In existing cache model, the cosine similarity between CLIP's image embeddings is computed to retrieve similar keys to a given query. Due to the above flaw of CLIP's image-image similarity, the ability of cache model would be limited. To mitigate the gap between pre-training and adaptation, we propose to calibrate the image-image similarity by exploiting some additional unlabeled images. We add some calibration layers on top of the CLIP's image encoder, and optimize their parameters using a self-supervised contrastive loss. To further improve the learning quality, we incorporate a hard mining strategy to construct more informative image batches. Once the calibration layers are trained, they can be seamlessly integrated into various downstream few-shot classification tasks. Fig.~\ref{fig:vis_sim_calib} visually demonstrates the effectiveness of our similarity calibration approach on the EuroSAT dataset~\citep{EuroSAT}. As evident from the figure, after calibration, the data points are more distinctly separated, indicating an improved ability to discern between visually similar and dissimilar images.

\textbf{Second}, existing cache model typically utilizes Gaussian kernel as the weight function to assess the significance of values stored in cache. However, without considering the normalization term, this form of cache model is a Nadaraya-Watson (N-W) estimator~\citep{NW}, which only considers the relation between query and keys, but neglects the relation among keys. The underlying problem of N-W estimator is that the weight function is positive anywhere, it cannot effectively capture negative correlations. Inspired by GP Regression, we introduce the precision matrix to reflect the relation among keys, effectively modeling negative correlations. By tuning the noise variance, N-W estimator will become a special case of the proposed method (as shown in Fig.~\ref{fig:weight_function}). As such, the GP-based cache model is more general and expressive.

\textbf{Third}, another crucial aspect overlooked by existing cache models is the predictive uncertainty. Conventionally, the output of these cache models is directly added to the logits of the zero-shot classifier, without considering the inherent uncertainty associated with the cache model. When a test sample is near to a training sample, the cache model's output should have a low uncertainty. In such a case, we should allocate a larger weight to cache model's output. In the opposite situation, a reduced weight should be assigned to account for higher uncertainty. The classical cache model, rooted in the N-W estimator, inherently struggles to quantify predictive uncertainty, as it lacks the mechanisms to generate an uncertainty value. However, by embracing the framework of GPs, we can naturally overcome this limitation. GPs provide not only a predictive mean but also a predictive variance, which serves as a direct measure of uncertainty. Higher variances correspond to greater uncertainties, enabling us to make informed decisions about the weight to assign to the cache model's predictions. Therefore, in our approach, we leverage the inverse of this predictive variance as the weighting factor for the cache model's output. In doing so, we enhance the reliability of our cache model by incorporating uncertainty-aware weighting, which is a natural and intuitive extension within the GP framework. 

Furthermore, to mitigate the computational complexity associated with GPs, we introduce a novel group-based learning strategy. This approach involves partitioning the classes into distinct groups and subsequently applying a separate GP model to each group. By disregarding the inter-group relationships and focusing on intra-group patterns, we achieve a substantial reduction in complexity. In summary, the key contributions of this work are listed as follows:

\begin{enumerate}
	\item We propose similarity calibration to bridge the gap between pre-training and adaptation of pre-trained contrastive vision-language models. 
	\item We propose the GP-based weight calibration to address the weakness of the existing cache model brought by N-W estimator. 
	\item We propose confidence calibration to introduce the predictive uncertainty of GP to logits computation. 
	\item We propose both training-free and training-required variant based on the above three calibration modules and propose a group-based learning strategy to reduce the complexity of GPs. We demonstrate the effectiveness of our method on 11 public datasets.
\end{enumerate}

\section{Related Work}
\label{sec:related_work}
\textbf{Vision-Language Models (VLMs)}. VLMs that simultaneously model the relationship between image and text modalities by leveraging a vast amount of paired image and text data available on the web, constitute a pivotal class of foundational model. According to the difference in learning task or objective, VLMs can be roughly categorized into two groups: generative VLMs~\citep{Flamingo,CoCa,VLBeit} typically employ the tasks of image captioning, masked language modeling (MLM), masked image modeling (MIM); while discriminative VLMs~\citep{Florence,CLIP,ALIGN} commonly use cross-modal contrastive learning as the objective. Compared to generative VLMs, discriminative VLMs are increasingly applied to various vision tasks, including zero-shot recognition~\citep{CLIP}, few-shot recognition~\citep{CoOp,cocoop}, image segmentation~\citep{CRIS}. At the heart of leveraging pre-trained VLMs for downstream tasks lies the art of parameter-efficient adaptation, which aims to fine-tune the models in a manner that minimizes the number of parameters need to be updated, thereby achieving efficient transfer learning while preserving the rich knowledge encoded in the pre-trained models. By carefully adapting VLMs in this manner, researchers and practitioners can harness their immense capabilities to tackle complex vision tasks with remarkable accuracy and efficiency.

\noindent\textbf{Parameter-efficient Adaptation of VLMs}. There are mainly two technical roadmaps for adapting VLMs in a parameter-efficient way. One line of work is inspired by the prompting technique~\citep{PrefixTuning,ScalePower,PromptSurvey} in NLP. Context optimization (CoOp)~\citep{CoOp} is the first attempt that introduces prompting to VLMs, which prepends learnable vectors as context to the inputted class name sequence of CLIP's text encoder and optimizes these extra parameters by gradient descent. Further improvements of CoOp include but not limited to enhancing the generalization ability~\citep{cocoop,ProGrad,KgCoOp,DingZYWXP24,NFL}, improving the accuracy~\citep{PDL,NPS}, integrating multi-task learning~\citep{SoftCPT}, introducing semi-supervised learning~\citep{ChakrabortySPD24}, exploring multi-modal and deeper prompts~\citep{MaPLe}. Despite the proven efficacy of prompt-based methods, they often necessitate substantial computational resources for learning these prompts.

Distinct from these approaches, recent research tends to design more efficient tuning methods by eliminating the demand of learning text prompts. CLIP-Adapter~\citep{CLIPAdapter}, for instance, conducted fine-tuning through feature adapters on either visual or language branch, demonstrating superior performance against CoOp while maintaining a simpler design. Following this trend, \cite{TipAdapter} proposed the training-free adaption method for CLIP, named Tip-Adapter, which constructs the adapter through a key-value based cache model from the few-shot training set and updates the prior knowledge contained in CLIP through feature retrieval. By enabling learnable keys, they also proposed a training-required variant named Tip-Adapter-F. Tip-Adapter and Tip-Adapter-F were demonstrated to be both effective and efficient. Based on Tip-Adapter, \cite{APE} proposed a training-free method APE and a training-required variant APE-T~\citep{APE} by decoupling domain-specific knowledge and exploring the trilateral affinities between test image, cache model and textual representation. Most recently, \cite{GDA} revisited the classical Gaussian Discriminant Analysis (GDA) algorithm and integrated it with the original zero-shot classifier, achieving state-of-the-art performance.

\noindent\textbf{Gaussian Processes (GPs)}. GPs are non-parametric supervised learning algorithm, and they could solve the regression and probabilistic classification problem. GPs exploit the GP prior that assumes any collection of function values adheres to a multivariate Gaussian distribution. Benefiting from the properties of Gaussian distribution, the posterior predictive distribution on new data points can be explicitly obtained. This work innovatively introduces the precision matrix from GP Regression into the cache model, thereby boosting its expressive ability substantially. Moreover, we introduce the estimated uncertainty in GPs to fused logits, mitigating the side-effect when test images deviate significantly from the training data.

A notable limitation of GPs lies in their computational intensiveness, primarily stemming from the necessity to invert large covariance matrix. To circumvent this obstacle, numerous approaches have been devised. For example, sparse GP~\citep{Titsias09} leverages a variational approach to identify a small set of inducing variables that can best approximate the posterior. Nystr{\"{o}}m Approximation~\citep{DrineasM05} also exploits a sampled subset of training samples and approximates the covariance matrix by the product of two low-rank matrices, each defined on this subset. Besides, Random Fourier Features (RFF)~\citep{RahimiR07} provides another avenue, mapping features to approximate Gaussian kernels via a randomized scheme. Different from these works, we reduce the complexity by a group-based learning strategy, which is inspired by the multi-task nature of GP Regression in this work. Compared to existing generic methods, we demonstrate that the proposed method is more suitable for the few-shot setting.

\noindent\textbf{Self-supervised Learning (SSL)}. SSL is a subset of unsupervised learning, which aims to learn feature representation using a vast amount of unlabeled data without human annotation. The paradigm of SSL is first learning universal feature representation based on a pretext task defined on unlabeled data and then transferring this representation to various downstream tasks by fine-tuning. The pretext tasks as the core of SSL include: context-based methods, contrastive learning and generative algorithms, etc. Typical context-based methods include rotation~\citep{GidarisSK18} and colorization~\citep{LarssonMS16}. Contrastive learning based methods are mainly relied on instance discrimination. MoCo v1~\citep{He0WXG20}, SimCLR v1~\citep{SimCLRv1} and SimCLR v2~\citep{SimCLRv2} are the typical representatives. MAE~\citep{MAE} and SimMIM~\citep{SimMIM} are based on masked image modeling task, which belong to the category of generative algorithms. This work identified the importance of calibrating the image-image similarity for cache model and explored the possibility of exploiting contrastive learning to address this issue. This could open a new door of combining SSL with vision-language model adaptation.

\section{Method}
\label{sec:method}

\subsection{Preliminary}
\label{ssec:pre}

\subsubsection{Gaussian Process Regression}
Instead of using GP Classification (GPC), we exploit the GP Regression (GPR) in our method, thus, this part only gives an introduction to GPR. For more information about GPs, please refer to the book of Rasmussen and Williams~\citep{RasmussenW06}.

Let $\mathcal{S}=\{\mathbf{x}_i, y_i\}_{i=1}^m$ denote a training set of $m$ i.i.d. samples from an unknown distribution. For notional convenience, we use $X=[\mathbf{x}_1,\cdots, \mathbf{x}_m]^T\in\mathbb{R}^{m\times d}$ and $\mathbf{y}=[y_1,\cdots, y_m]^T\in\mathbb{R}^m$ to collect all the input variables and the corresponding response values, where $d$ is the dimension of input space. The GPR model builds the input and output relation by the following equation:
\begin{align}
y_i=f(\mathbf{x}_i) + \epsilon_i, \; i=1,\cdots, m,
\end{align}
where $\epsilon_i,i=1,\cdots, m$ denote noise and obey independent $\mathcal{N}(0, \sigma^2)$ with $\sigma^2$ the noise variance. $f(\cdot)$ denotes a predictive function and is assumed to have a zero-mean GP prior:
\begin{align}
f(\cdot)\sim \mathcal{GP}(0, \kappa(\cdot, \cdot)),
\end{align}
where $\kappa(\cdot, \cdot)$ is a valid covariance function.

Let $\mathcal{T}=\{\mathbf{x}_i^*, y_i^*\}_{i=1}^{m^*}$ denotes a testing set of $m^*$ i.i.d. samples from the same distribution as the training set. Similarly, let us denote $X^*=[\mathbf{x}_1^*,\cdots, \mathbf{x}^*_m]^T\in\mathbb{R}^{m^*\times d}$ and $\mathbf{y}^*=[y_1^*,\cdots, y_m^*]^T\in\mathbb{R}^{m^*}$. Note that, for GPs, any finite samples of functions are jointly Gaussian distributed. As such, we have
\begin{align}
\left.
\begin{bmatrix}
\mathbf{f}\\\mathbf{f}^*
\end{bmatrix}
\right\vert
X, X^* \sim\mathcal{N}\left(\mathbf{0}, \begin{bmatrix}K(X, X)&K(X, X^*)\\K(X^*, X)&K(X^*, X^*)\end{bmatrix}\right),
\end{align}
where $\mathbf{0}$ denotes a all-zero vector, $\mathbf{f}=[f(\mathbf{x}_1), \cdots, f(\mathbf{x}_{m})]^T$ and $\mathbf{f}^*=[f(\mathbf{x}_1^*), \cdots, f(\mathbf{x}_{m^*}^*)]^T$. $K(X, X)\in\mathbb{R}^{m\times m}$, $K(X, X^*)\in\mathbb{R}^{m\times m^*}$, $K(X^*, X)\in\mathbb{R}^{m^*\times m}$ and $K(X^*, X^*)\in\mathbb{R}^{m^*\times m^*}$ are covariance matrices, which are defined as $(K(X, X))_{ij}=\kappa(\mathbf{x}_i, \mathbf{x}_j)$, $(K(X, X^*))_{ij}=\kappa(\mathbf{x}_i, \mathbf{x}^*_j)$, $(K(X^*, X))_{ij}=\kappa(\mathbf{x}^*_i, \mathbf{x}_j)$, and $(K(X^*, X^*))_{ij}=\kappa(\mathbf{x}^*_i, \mathbf{x}^*_j)$, respectively.

With the assumption of i.i.d. noise, the joint probability distribution of $\mathbf{y}$ and $\mathbf{y}^*=[y_1^*,\cdots, y^*_{m^*}]$ is
\begin{align}
\!\!
\left.
\begin{bmatrix}
\mathbf{y}\\\mathbf{y}^*
\end{bmatrix}
\right\vert
X, X^*
\!\sim\!\mathcal{N}\left(\mathbf{0}, \!\begin{bmatrix}K(X, \!X)\!+\!\sigma^2\mathbb{I}&K(X, \!X^*)\\K(X^*, \!X)&K(X^*, \!X^*)\end{bmatrix}\right),\!\!
\end{align}
where $\mathbb{I}$ denotes the identity matrix. With the rule for conditioning Gaussian, the posterior distribution of $\mathbf{y}^*$ is still a Gaussian:
\begin{align}
\mathbf{y}^*|\mathbf{y},X,X^*\sim \mathcal{N}(\mu^*, \Sigma^*),
\end{align}
where
\begin{align}
\mu^*&\!=\!K(X^*,X)(K(X,X)+\sigma^2 \mathbb{I})^{-1}\mathbf{y},\label{eq:gp_mean}\\
\Sigma^*&\!=\!K(X^*,X^*)\!-\!K(X^*,X)(K(X,X)\!+\!\sigma^2 \mathbb{I})^{-1}K(X, X^*).\label{eq:gp_var}
\end{align}
In Eq.~\ref{eq:gp_mean}, $\mu^*$ is the mean prediction, which could be viewed as a linear combination of observations $\mathbf{y}$ with the weight function $K(X^*,X)(K(X,X)+\sigma^2\mathbb{I})^{-1}$. In Eq.~\ref{eq:gp_var}, $\Sigma^*$ is the predicted covariance matrix, whose diagonal element denotes the predictive variance for each testing sample. A merit of the variance is that it indicates the uncertainty of the regression model. Smaller variance implies higher certainty.

\begin{figure*}[t!]
	\centering
	\includegraphics[width=0.89\linewidth]{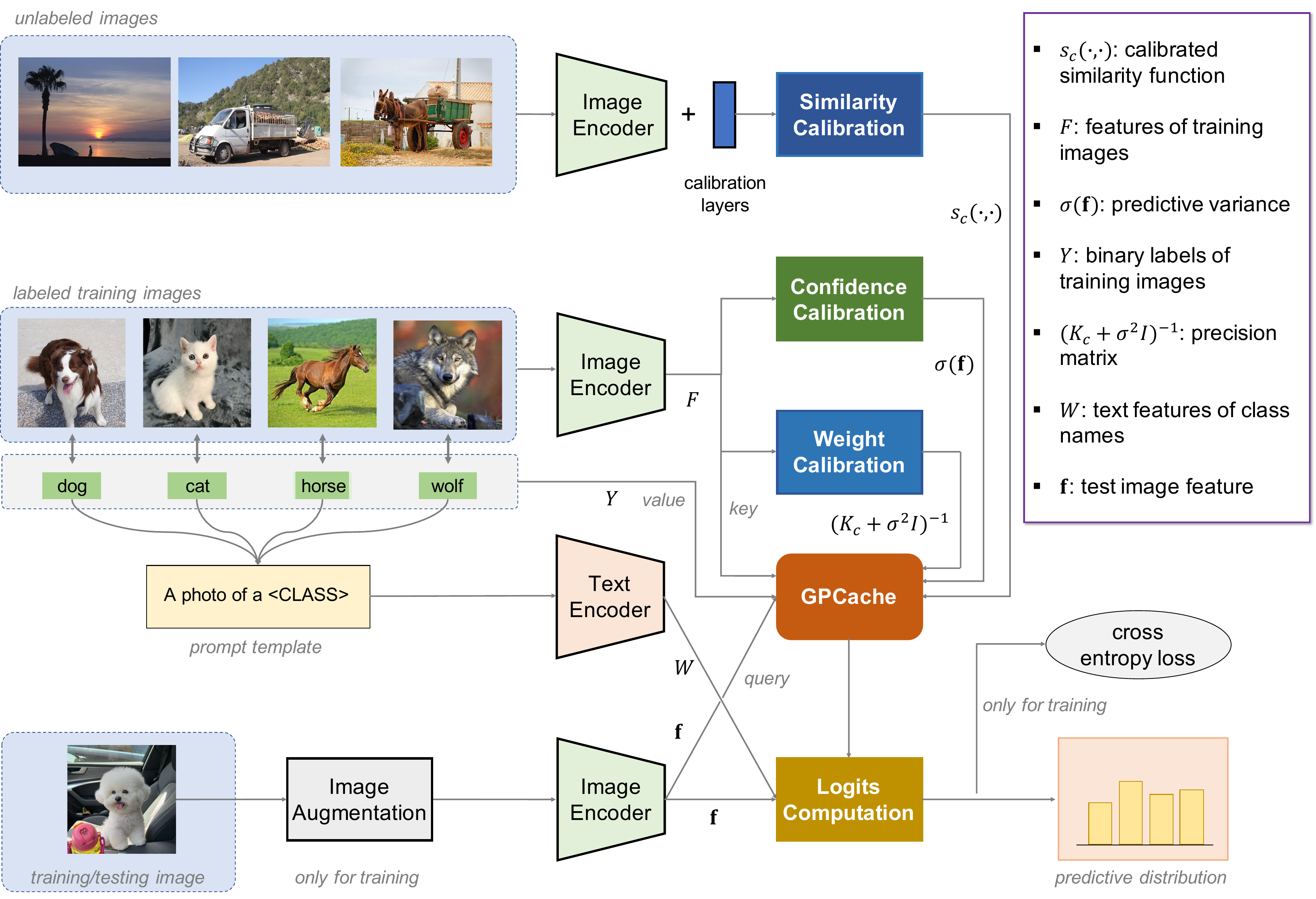}
	\caption{\textbf{Overview of the proposed adaptation method GPCache for vision-language models (VLMs).} The key ingredients of the proposed method are similarity calibration, confidence calibration and weight calibration.}
	\label{fig:overview}
	\vspace{-3pt}
\end{figure*}

\subsubsection{Cache Model}
We also introduce the cache model in existing efficient tuning methods, such as Tip-Adapter~\citep{TipAdapter} and APE~\citep{APE}. Cache based methods for few-shot adaptation combine two logits components, one is the logits generated by zero-shot classifier, and the other is from cache model. 

Let us assume that there are $n$ labeled training images in training set, i.e., $(I_i, \mathbf{y}_i), i=1,\cdots, n$, where $I_i$ and $\mathbf{y}_i\in\mathbb{R}^c$ denote the $i$-th training image and the corresponding one-hot binary label vector, respectively. $c$ denotes the number of classes. We extract the image features of each image by CLIP's image encoder denoted as $f_{img}()$, i.e., $\mathbf{f}_i=f_{img}(I_i),i=1,\cdots, n$. Assume CLIP's text encoder is $f_{txt}()$, we can obtain the text features corresponding to each class: $\mathbf{w}_i=f_{txt}(\text{``A photo of a } T_i")$, where $T_i$ is the class name of the $i$-th class. For convenience, we also assume that $\mathbf{f}_i$ and $\mathbf{w}_i$ are already L2-normalized.

Given a test image $I$, we extract its feature by $f_{img}()$ and obtain $\mathbf{f}=f_{img}(I)$. The logits given by zero-shot classifier are computed by
\begin{align}
\text{logits}_{zs}=[\mathbf{f}^T\mathbf{w}_1, \cdots, \mathbf{f}^T\mathbf{w}_c]=\mathbf{f}^T\cdot W,
\end{align}
where $W=[\mathbf{w}_1,\cdots, \mathbf{w}_c]$.

The cache model consists of key-value pairs. The keys are the image features $\mathbf{f}_i=f_{img}(I_i),i=1,\cdots, n$ and the values are their corresponding binary label vectors $\mathbf{y}_i,i=1,\cdots, n$. The logits outputted by cache model are computed by the following equation:
\begin{align}
\text{logits}_{cc}=\sum\nolimits_{i=1}^n \kappa(\mathbf{f}, \mathbf{f}_i)\cdot \mathbf{y}_i^T,\label{eq:cache_logits}
\end{align}
where $\kappa(\mathbf{f}_i, \mathbf{f}_j)$ is the Gaussian kernel function: $\kappa(\mathbf{f}_i, \mathbf{f}_j)=\exp(-\beta(1-\mathbf{f}^T_i\mathbf{f}_j))$ with $\sqrt{2/\beta}$ the bandwidth parameter.

The final logits in Tip-Adapter is the sum of $\text{logits}_{zs}$ and $\text{logits}_{cc}$, i.e., 
\begin{align}
\text{logits}=\text{logits}_{zs}+\alpha\cdot \text{logits}_{cc},\label{eq:tip_logits}
\end{align}
where $\alpha$ is a weight coefficient for tuning the relative importance of cache model and zero-shot prediction. As for APE, the term $\text{logits}_{cc}$ is further modified by decoupling domain-specific knowledge.

\subsection{GPCache}
\label{ssec:gpcache}
The overall framework of the proposed method is shown in Fig.~\ref{fig:overview}. As can be seen, the proposed method for adapting VLMs takes a new cache model GPCache as its core, which is based on three calibration modules: similarity calibration, confidence calibration and weight calibration. The similarity calibration module optimizes the image-image similarity function on unlabeled images. The confidence calibration module takes training images' features as input and computes the cache model's predictive variance, which is used to re-scale the logits generated by cache model. The weight calibration module computes the precision matrix which is used to tune the weight function in cache model. For a test image, the logits computation module sums up the logits generated by zero-shot classifier and the logits by GPCache. The fused logits are used for classification. For training-free mode, there are no learnable parameters; while for training-required mode, we allow the keys to be tunable and they are learned by minimizing cross entropy loss. In the following sections, we will detail the main ingredients of the proposed method.

\begin{figure*}[t!]
	\centering
	\includegraphics[width=0.9\linewidth]{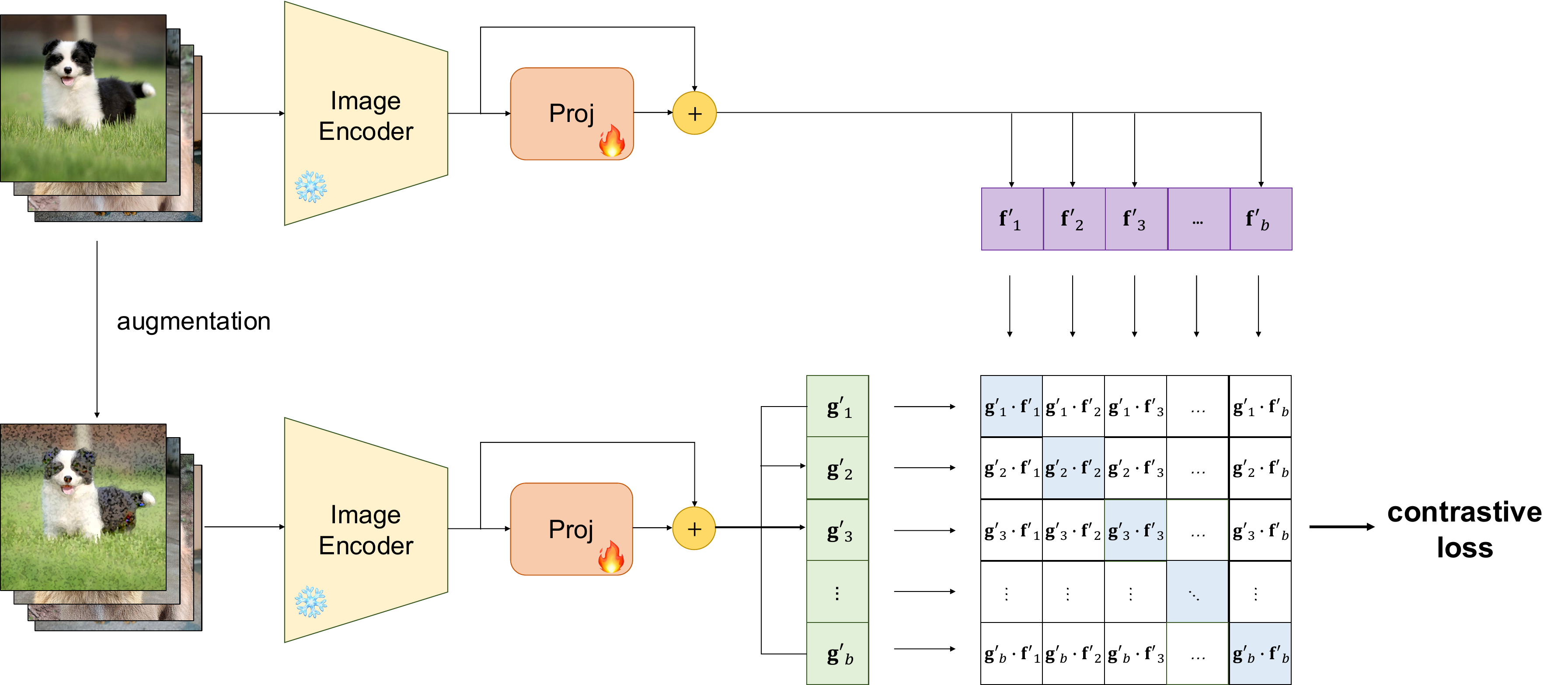}
	\caption{\textbf{Illustration of computing contrastive loss in the similarity calibration stage. }$\mathbf{f}_1,\cdots, \mathbf{f}_b$ are extracted features of original images, $\mathbf{g}_1,\cdots, \mathbf{g}_b$ are extracted features of augmented images.}
	\label{fig:sim_calib}
	\vspace{-5pt}
\end{figure*}

\subsubsection{Similarity Calibration} We propose to calibrate the image-image similarity using unlabeled images. To achieve this goal, we introduce learnable calibration layers on top of CLIP's image encoder $f_{img}()$. To avoid forgetting pre-trained knowledge, we exploit a residual structure. The calibration layers consist of a projection layer, a residual connection and a normalization layer, which are denoted as:
\begin{align}
\phi(\mathbf{f}; \theta)=\text{Norm}(\text{Proj}(\mathbf{f}; \theta) + \mathbf{f}),
\end{align}
where $\theta$ collects the learnable parameters, $\mathbf{f}=f_{img}(I)$ denotes the feature of some image $I$, $\text{Norm}(\cdot)$ denotes the L2 normalization operator and $\text{Proj}(\cdot)$ signifies the projection layer. We find that a simple linear projection layer works well.

To learn the parameters $\theta$, we assume there are $p$ images available, which are denoted as $I_i, i=1,\cdots,p$. For each image, we extract its feature by $f_{img}()$, which is denoted as $\mathbf{f}_i$. The corresponding feature after applying the calibration layers is denoted as $\mathbf{f}'_i=\phi(\mathbf{f}_i; \theta)$. Different from CLIP, we define intra-modal contrastive loss for parameter optimization. The self-supervised learning helps to learn similarity with good generalization ability compared to supervised learning. Fig.~\ref{fig:sim_calib} gives an illustrative figure of the intra-modal contrastive loss. 

For self-supervised training, we randomly augment the images $I_i, i=1,\cdots,p$, and compute their calibrated features, which are denoted as ${\mathbf{g}}'_i, i=1,\cdots,p$. For mini-batch based optimization, we first randomly sample a batch of $b$ images from the unlabeled training set. For each image in this batch, we find $r$ most similar images from the unlabeled training set to it and define the contrastive loss on these $r$ images. Finally, the overall loss on this batch is defined as
\begin{align}
\mathcal{L}_{cl}(\theta) =\frac{1}{b} \sum\limits_{t=1}^b \sum\limits_{i\in \mathcal{I}_t} \log \frac{\exp(z_{ii}/\tau)}{\sum\nolimits_{j\in \mathcal{I}_t} \exp(z_{ij}/\tau)},
\end{align}
where $t$ is the image index within this batch, $\mathcal{I}_t$ is the index set of $r$ similar images within the whole training set retrieved for the $t$-th image of the batch. Note that, the retrieval is based on cosine similarity of original image features. Besides, $\tau$ is the temperature parameter and $z_{ij}$ denotes the image-image similarity. In this work, we find that $z_{ij}=(\mathbf{g}'_i)^T\mathbf{f}'_j$ works well. The above contrastive loss differs to existing loss in SimCLR~\citep{SimCLRv1}, since we perform hard mining by nearest searching. In the experiment, we will demonstrate hard mining helps to learn better similarity function.

By using gradient descent algorithm to minimize $\mathcal{L}_{cl}(\theta)$, we obtain the optimal parameter $\theta^*$. As such, the calibrated image-image similarity is expressed as:
\begin{align}
s_{c}(\mathbf{f}_i, \mathbf{f}_j)=\phi(\mathbf{f}_i;\theta^*)^T\phi(\mathbf{f}_j;\theta^*),
\end{align}
where $\mathbf{f}_i$ and $\mathbf{f}_j$ are image features. By contrast, the original image-image similarity without calibration is
\begin{align}
s(\mathbf{f}_i, \mathbf{f}_j)=\mathbf{f}_i^T\mathbf{f}_j.
\end{align}

\subsubsection{Weight Calibration} The logits by cache model is given in Eq.~\ref{eq:cache_logits}, which can be reformulated into
\begin{align}
\text{logits}_{cc}=\kappa(\mathbf{f}, F) \cdot Y,
\end{align}
where $Y=[\mathbf{y}_1, \cdots, \mathbf{y}_n]^T\in\mathbb{R}^{n\times c}$ is the binary label matrix and $\kappa(\mathbf{f}, F)=[\kappa(\mathbf{f}, \mathbf{f}_i),\cdots, \kappa(\mathbf{f}, \mathbf{f}_n)]$ collects all the weights. Without considering the normalization term, this equation is actually a N-W estimator. As mentioned above, the weights omit the relation between training samples. Inspired by GPs, we introduce the precision matrix into the cache model:
\begin{align}
\text{logits}'_{cc}=\kappa_c(\mathbf{f}, F) (K_c+\sigma^2\mathbb{I})^{-1} \cdot Y, 
\end{align}
where $(K_c+\sigma^2\mathbb{I})^{-1}$ is the precision matrix, $K_c=[\kappa_c(\mathbf{f}_i, \mathbf{f}_j)]_{ij=1}^n$ is the covariance matrix and $\kappa_c(\mathbf{f}, F)=[\kappa_c(\mathbf{f}, \mathbf{f}_i),\cdots, \kappa_c(\mathbf{f}, \mathbf{f}_n)]$. $\kappa_c(\cdot, \cdot)$ is the Gaussian kernel with calibrated image-image similarity function:
\begin{align}
\kappa_c(\mathbf{f}_i, \mathbf{f}_j)=\exp(-\beta (1-s_{c}(\mathbf{f}_i, \mathbf{f}_j))).
\end{align}

\subsubsection{Confidence Calibration} In Eq.~\ref{eq:cache_logits}, when the test sample is far from all training samples, the logits by the cache model will become $\text{logits}_{cc}=\frac{1}{n}\sum\nolimits_{i=1}^n \mathbf{y}_i^T=\frac{k}{n}\mathbf{1}^T$ with $k$ the number of training samples per class, which provides no useful information. When the test sample approaches to a certain training sample, the logits become $\text{logits}_{cc}\approx \mathbf{y}_t^T$ ($t\in [1, n]$) with sufficiently large $\beta$. In such a case, the logits by cache model are very informative. As such, it is important to use different weight for different test sample. GPR belongs to the Bayesian learning framework, which can naturally provide uncertainty information. In this work, we exploit the predictive variance of GPR to dynamically tune the importance of cache model.

For a single test image with feature denoted as $\mathbf{f}$, the GPR variance can be expressed as
\begin{align}
\sigma(\mathbf{f})^2=1-\kappa_c(\mathbf{f},F)(K_c+\sigma^2\mathbb{I})^{-1}\kappa_c(\mathbf{f},F)^T.
\end{align}
We re-scale the logits $\text{logits}'_{cc}$ by the inverse of $\sigma(\mathbf{f})$ and obtain:
\begin{align}
\text{logits}''_{cc}=\frac{\kappa_c(\mathbf{f}, F) (K_c+\sigma^2\mathbb{I})^{-1} \cdot Y}{\left[1-\kappa_c(\mathbf{f},F)(K_c+\sigma^2\mathbb{I})^{-1}\kappa_c(\mathbf{f},F)^T\right]^{\eta}},
\end{align}
which is the final logits outputted by GPCache. Here, $\eta$ is introduced to control the numerical distribution of weights.

\subsubsection{Logits Computation} Similar to existing methods, the final logits for classification is the sum of the logits given by the zero-shot classifier and the GPCache, that is
\begin{align}
\text{logits}=\text{logits}_{zs}+\alpha\cdot \text{logits}''_{cc}.\label{eq:final_logits}
\end{align}
Note that, different from $\sigma(\mathbf{f})^2$, the parameter $\alpha$ is a global scaling factor. 

\subsubsection{Training-required Extension} Based on the logits in Eq.~\ref{eq:final_logits}, we can already make prediction for a given test image. However, in order to fully utilize the information from training set, we further propose a training-required variant. Here, we unfreeze the keys in GPCache, i.e., $F$, and optimize it by minimizing the cross entropy loss on training set. Note that, the initial value of $F$ is given by the pre-trained image encoder of CLIP and we also freeze the parameters in the calibration layers. We name the training-free and training-required method as GPCache and GPCache-F, respectively.

\subsubsection{Reducing Complexity} As we need to invert the matrix $K_c+\sigma^2\mathbb{I}$, the storage and computational complexity of GPCache are high, i.e., $\mathcal{O}(c^2k^2)$ and $\mathcal{O}(c^3k^3)$, respectively. In the few-shot setting, $k$ is quite small, however, $c$ could be relatively large. As such, directly applying GPCache to dataset of many classes could be questionable. When there are many classes, existing general kernel approximation methods like RFF will encounter problems, as the sensitivity to approximation error will be enlarged with increased number of classes. For training-free methods, the side impact of approximation error is even greater as it could not be diminished by further tuning like training-required methods. Instead of using existing methods for reducing complexity, we propose a group-based learning strategy which exploits the multi-task learning nature of GP in GPCache.

Note that there are $c$ classes in our setting, which correspond to $c$ univariate GP regressors. For each regression task, all of the $n$ training samples are adopted. Actually, for each regression task, we only need to sample a subset of the samples. Specifically, we randomly split the $c$ classes into $g$ groups and fit one GP regressor for each group only with the training samples from the corresponding $c/g$ classes. Assume that the training feature matrix of the $i$-th group is $F_i$ and the associated binary label matrix is $Y_i\in\mathbb{R}^{n\times c/g}$, the logits of this group can be represented as
\begin{align}
\text{logits}''_{cc}(i)\!=\!\frac{\kappa_c(\mathbf{f}, F_i) (K_c^i+\sigma^2\mathbb{I})^{-1} \cdot Y_i}{\left[1\!-\!\kappa_c(\mathbf{f},F_i)(K_c^i\!+\!\sigma^2\mathbb{I})^{-1}\kappa_c(\mathbf{f},F_i)^T\right]^{\eta}},
\end{align}
where $K_c^i$ is the covariance matrix of the $i$-th group. The final logits of GPCache are the concatenation of logits from all groups:
\begin{align}
\text{logits}''_{cc}=[\text{logits}''_{cc}(1), \cdots, \text{logits}''_{cc}(g)].
\end{align}
Based on the proposed group-based learning strategy, the storage and computational complexity are reduced to $\mathcal{O}({c^2k^2}/{g})$ and $\mathcal{O}({c^3k^3}/{g^2})$, respectively.

\section{Experiments}
\label{sec:exp}

\subsection{Setup}

\textbf{Dataset.} Following previous arts~\citep{TipAdapter,CoOp,APE}, we adopt the commonly used 11+2 public image classification datasets to evaluate the effectiveness of our proposed method under two settings: \textbf{few-shot classification} and \textbf{domain generalization}. The datasets for few-shot recognition cover a wide range of different recognition tasks, including generic object classification (ImageNet~\citep{ImageNet} and Caltech101~\citep{Caltech101}), fine-grained image classification (Food101~\citep{Food101}, OxfordPets~\citep{OxfordPets}, StanfordCars~\citep{StanfordCars},  Flowers102~\citep{Flowers102} and FGVCAircraft~\citep{FGVCAircraft}), action recognition (UCF101~\citep{UCF101}), satellite image classification (EuroSAT~\citep{EuroSAT}), scene recognition (SUN397~\citep{SUN397}), texture classification (DTD~\citep{DTD}). As for domain generalization setting, we use ImageNet as the source domain, and use ImageNetV2~\citep{ImageNetV2} and ImageNet-Sketch~\citep{ImageNetS} as the target domains.

\begin{figure*}[!htbp]
	\centering
	\begin{minipage}{0.24\textwidth}
		\centering
		\includegraphics[width=1.0\textwidth]{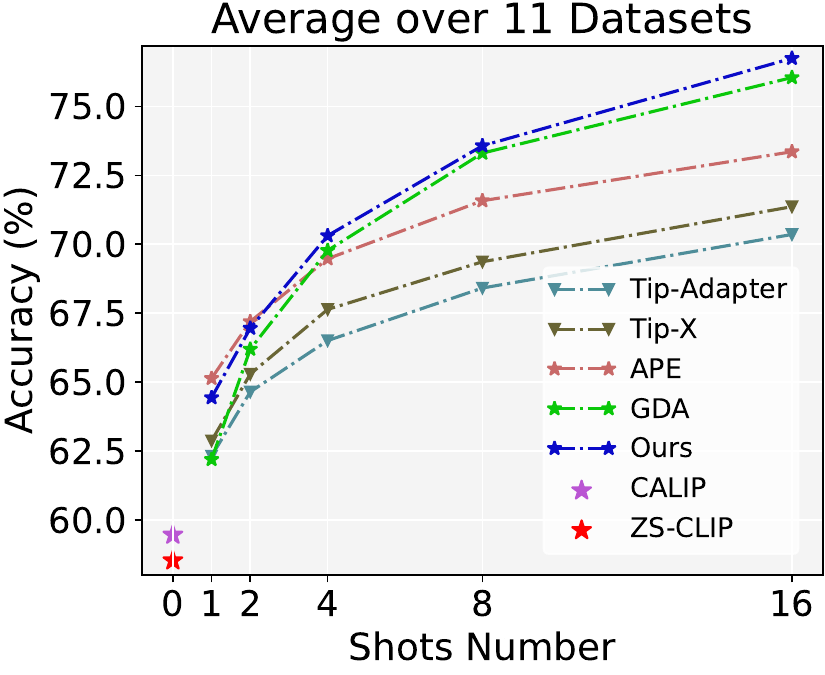}
	\end{minipage}
	\begin{minipage}{0.24\textwidth}
		\centering
		\includegraphics[width=1.0\textwidth]{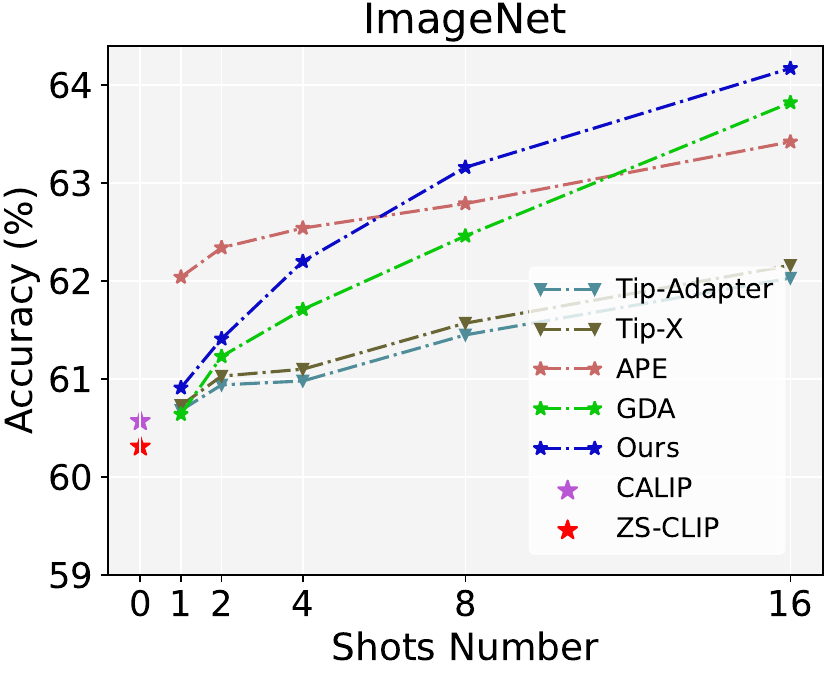}
	\end{minipage}
	\begin{minipage}{0.24\textwidth}
		\centering
		\includegraphics[width=1.0\textwidth]{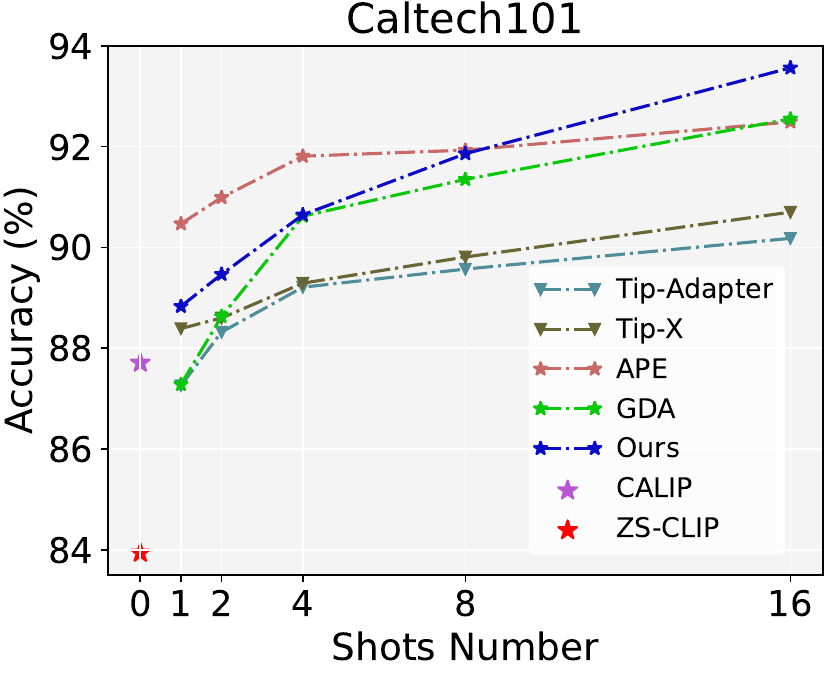}
	\end{minipage}
	\begin{minipage}{0.24\textwidth}
		\centering
		\includegraphics[width=1.0\textwidth]{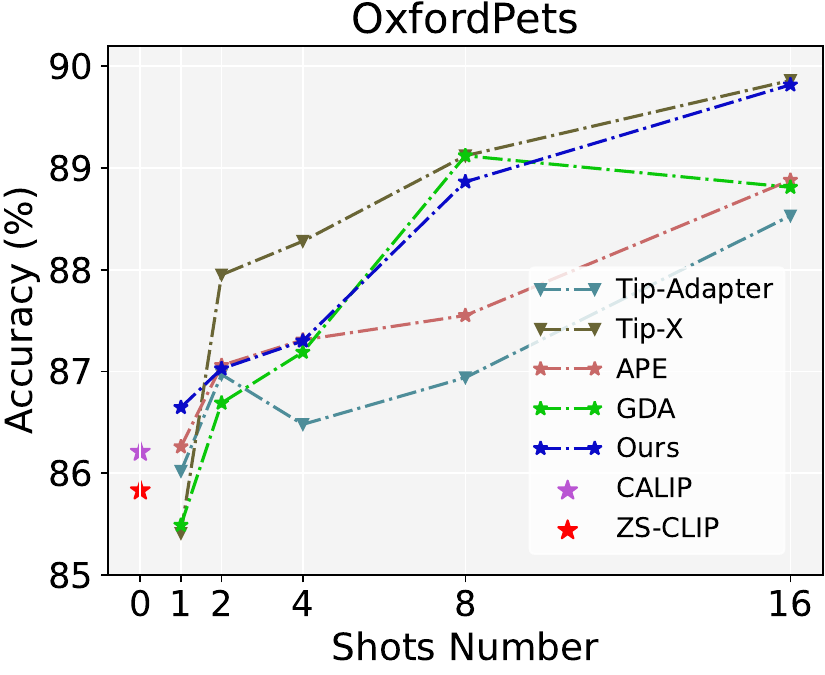}
	\end{minipage}
\\
	\begin{minipage}{0.24\textwidth}
		\centering
		\includegraphics[width=1.0\textwidth]{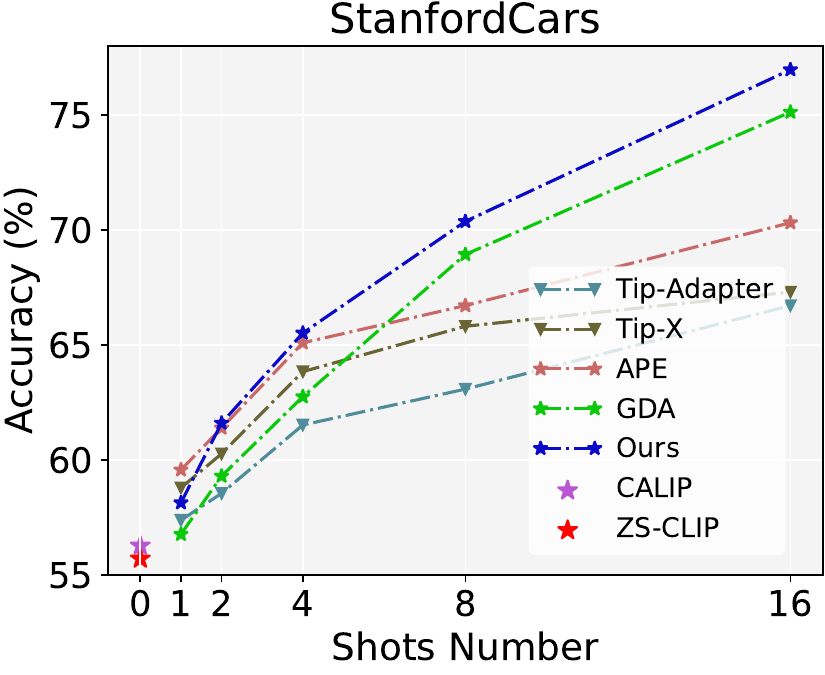}
	\end{minipage}
	\begin{minipage}{0.24\textwidth}
		\centering
		\includegraphics[width=1.0\textwidth]{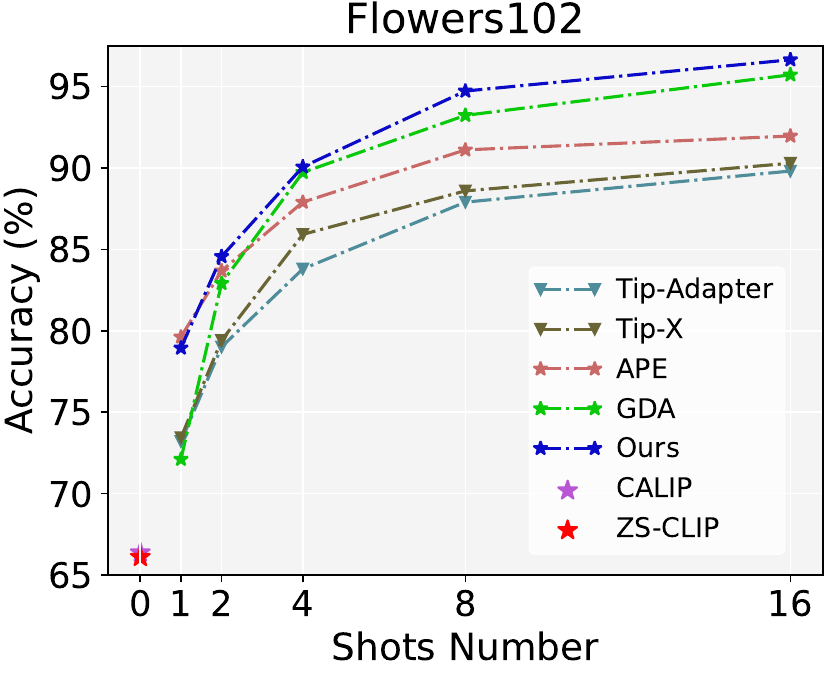}
	\end{minipage}
	\begin{minipage}{0.24\textwidth}
		\centering
		\includegraphics[width=1.0\textwidth]{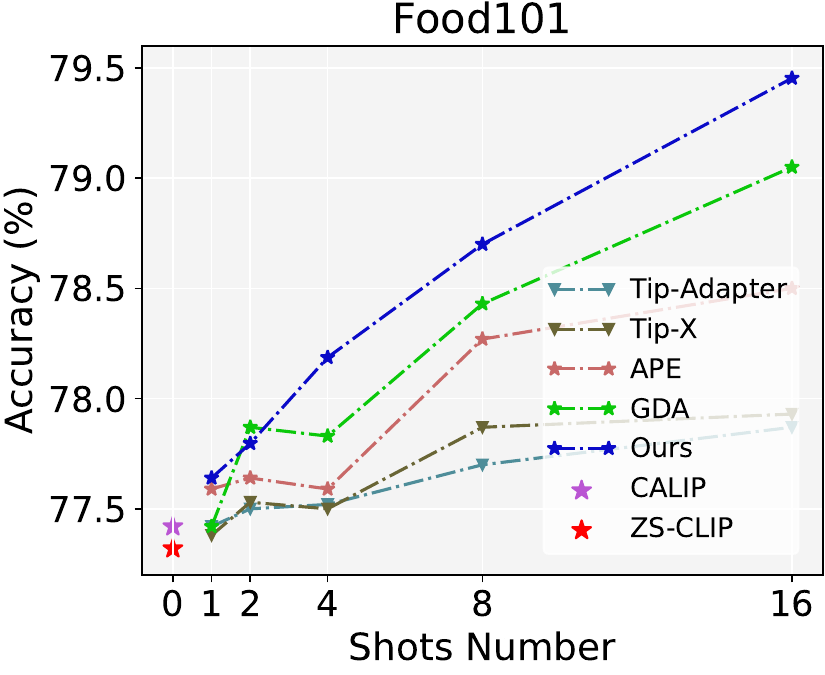}
	\end{minipage}
	\begin{minipage}{0.24\textwidth}
		\centering
		\includegraphics[width=1.0\textwidth]{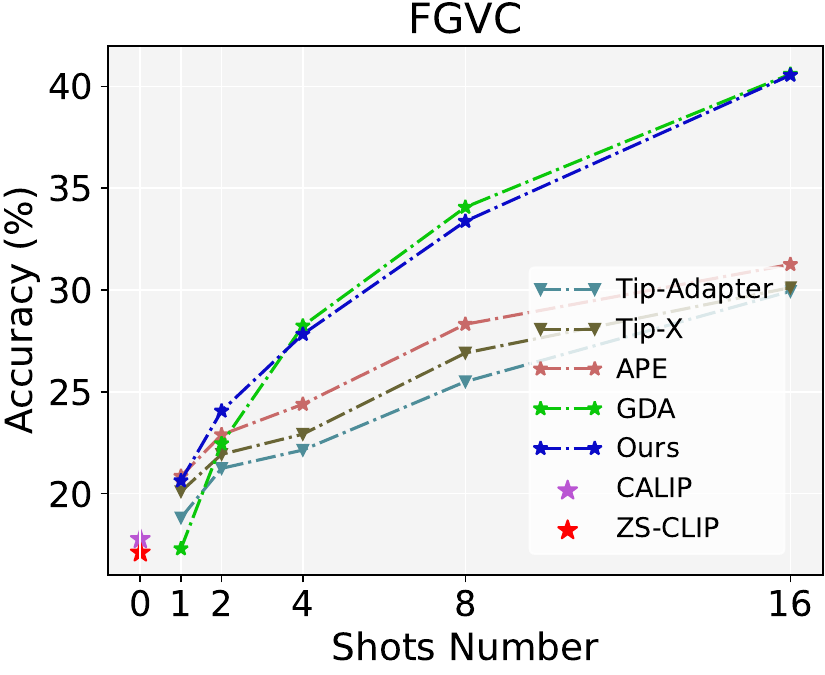}
	\end{minipage}
\\
	\begin{minipage}{0.24\textwidth}
		\centering
		\includegraphics[width=1.0\textwidth]{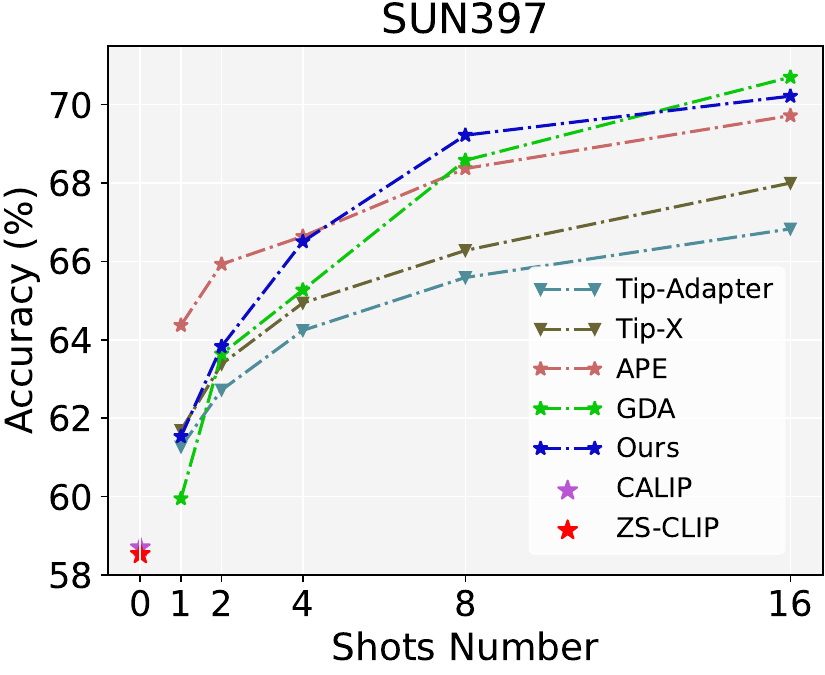}
	\end{minipage}
	\begin{minipage}{0.24\textwidth}
		\centering
		\includegraphics[width=1.0\textwidth]{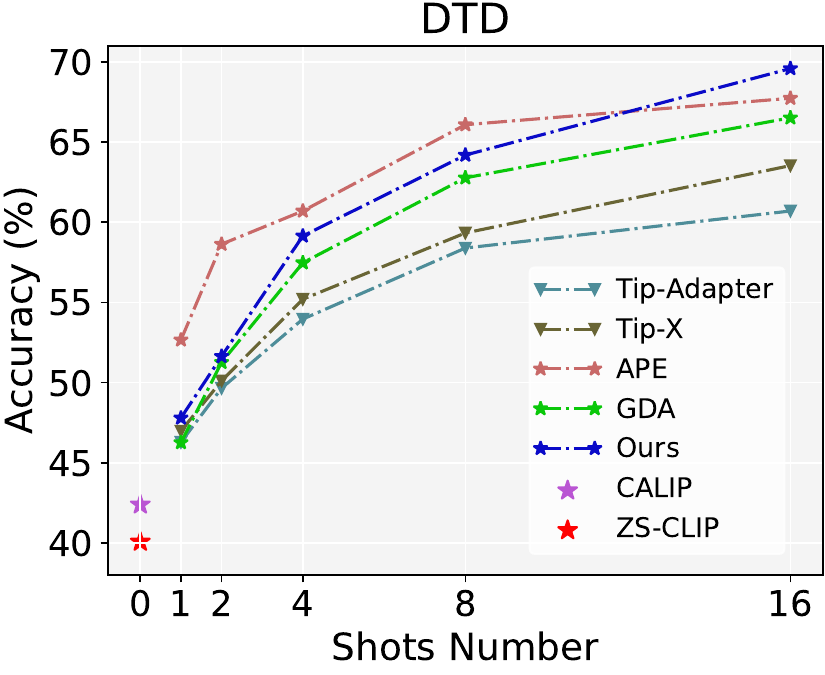}
	\end{minipage}
	\begin{minipage}{0.24\textwidth}
		\centering
		\includegraphics[width=1.0\textwidth]{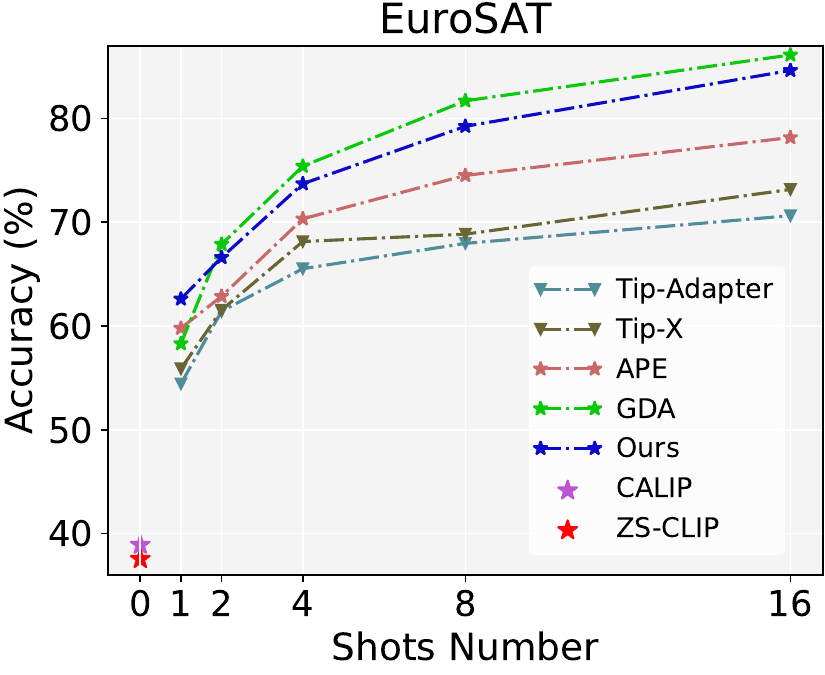}
	\end{minipage}
	\begin{minipage}{0.24\textwidth}
		\centering
		\includegraphics[width=1.0\textwidth]{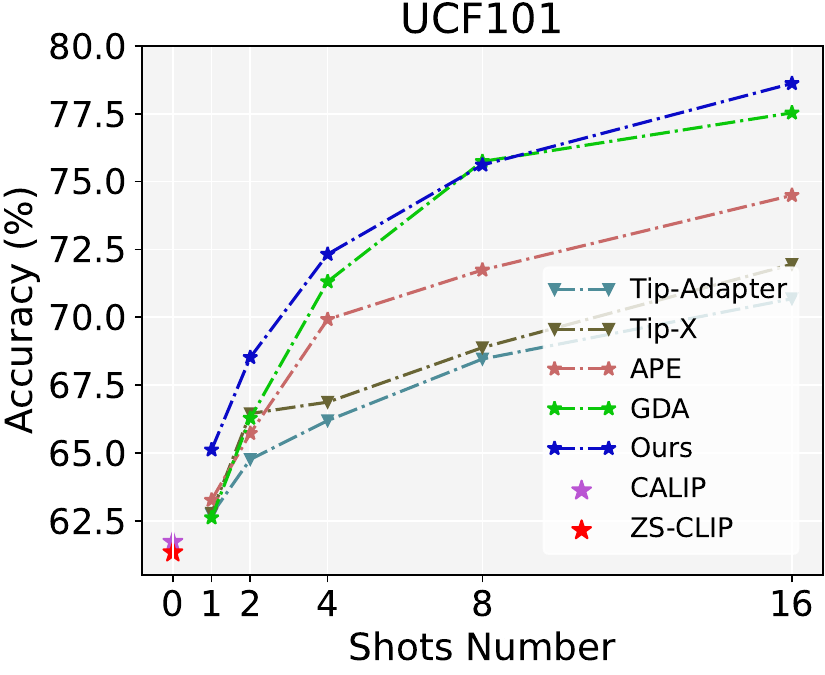}
	\end{minipage}
	\caption{
		{Comparison of training-free methods under the few-shot classification setting on the 11 datasets.}
	}
	\vspace{-10pt}
	\label{fig:trainfree_fewshots}
\end{figure*}

\begin{figure*}[!htbp]
	\centering
	\begin{minipage}{0.24\textwidth}
		\centering
		\includegraphics[width=1.0\textwidth]{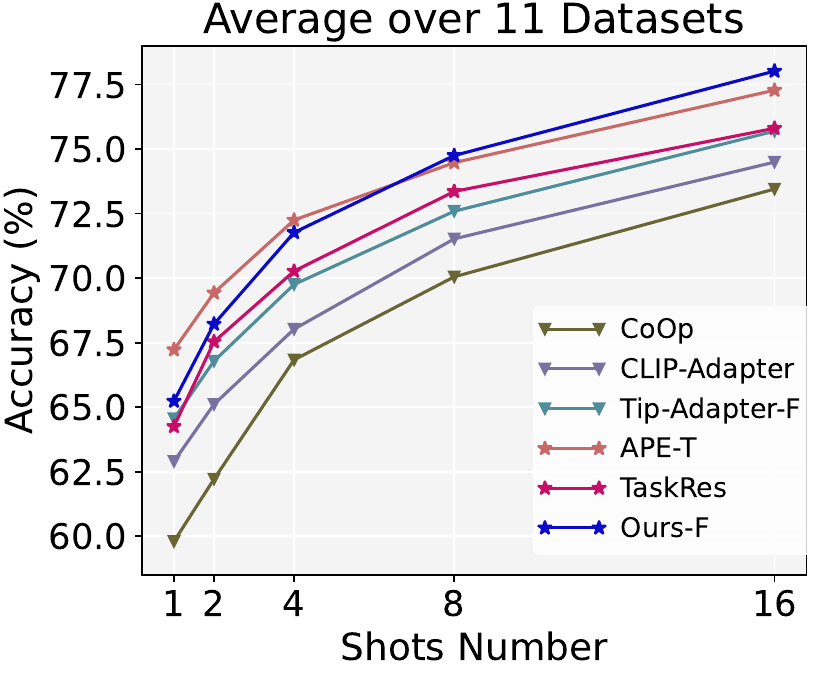}
	\end{minipage}
	\begin{minipage}{0.24\textwidth}
		\centering
		\includegraphics[width=1.0\textwidth]{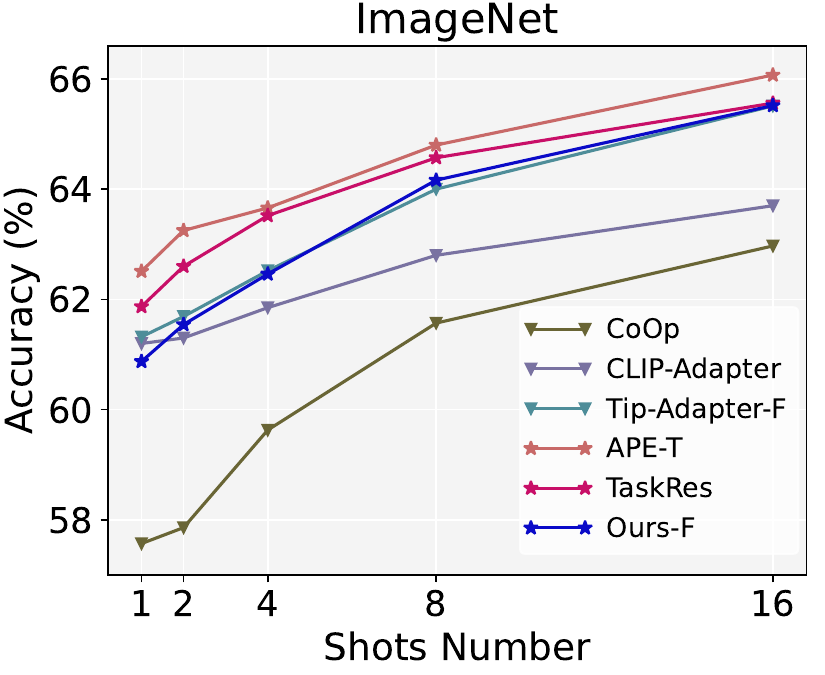}
	\end{minipage}
	\begin{minipage}{0.24\textwidth}
		\centering
		\includegraphics[width=1.0\textwidth]{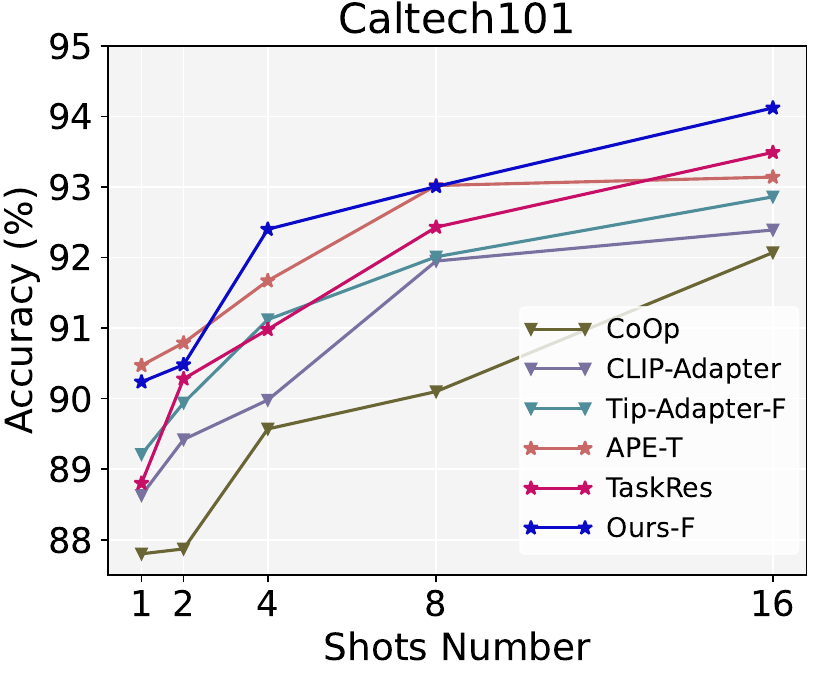}
	\end{minipage}
	\begin{minipage}{0.24\textwidth}
		\centering
		\includegraphics[width=1.0\textwidth]{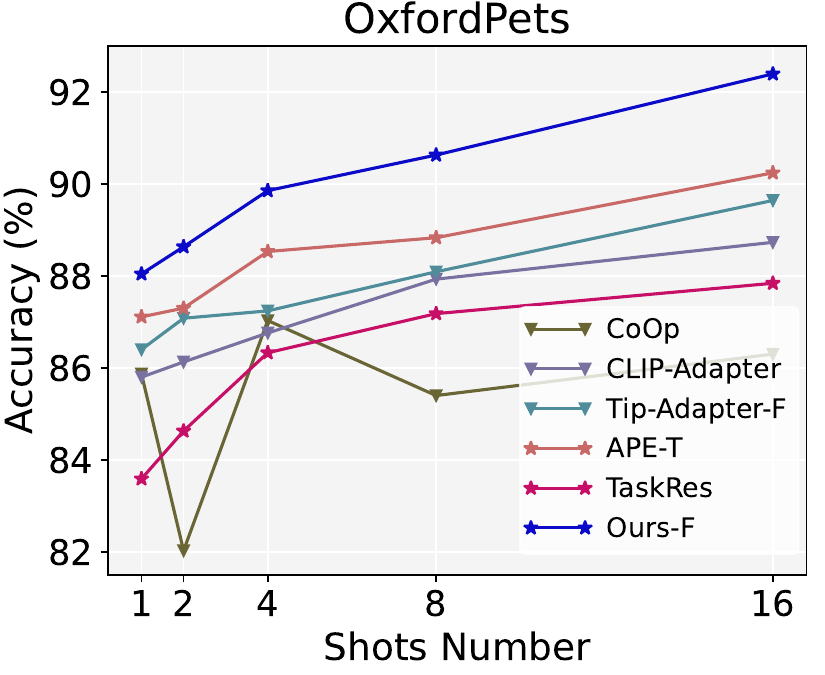}
	\end{minipage}
\\
	\begin{minipage}{0.24\textwidth}
		\centering
		\includegraphics[width=1.0\textwidth]{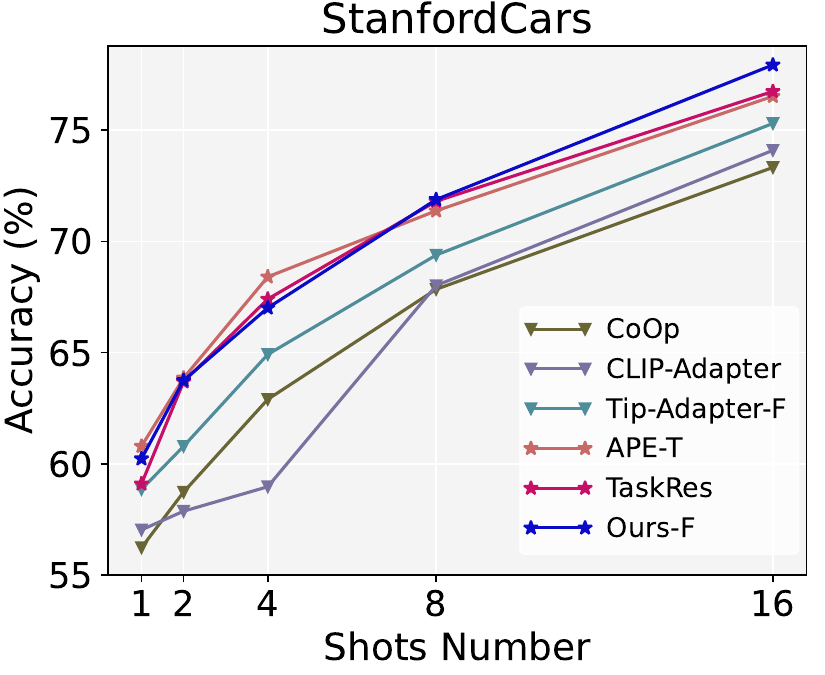}
	\end{minipage}
	\begin{minipage}{0.24\textwidth}
		\centering
		\includegraphics[width=1.0\textwidth]{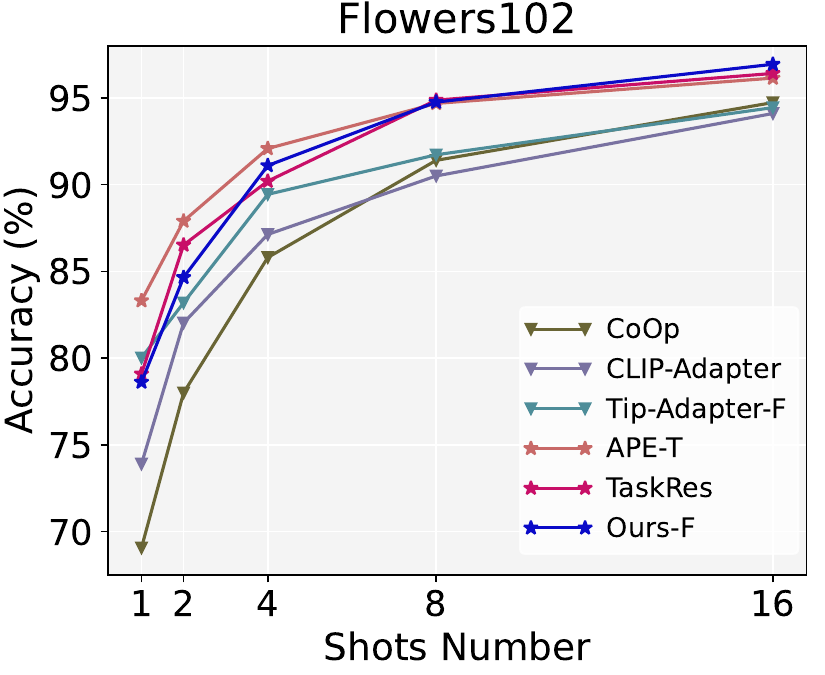}
	\end{minipage}
	\begin{minipage}{0.24\textwidth}
		\centering
		\includegraphics[width=1.0\textwidth]{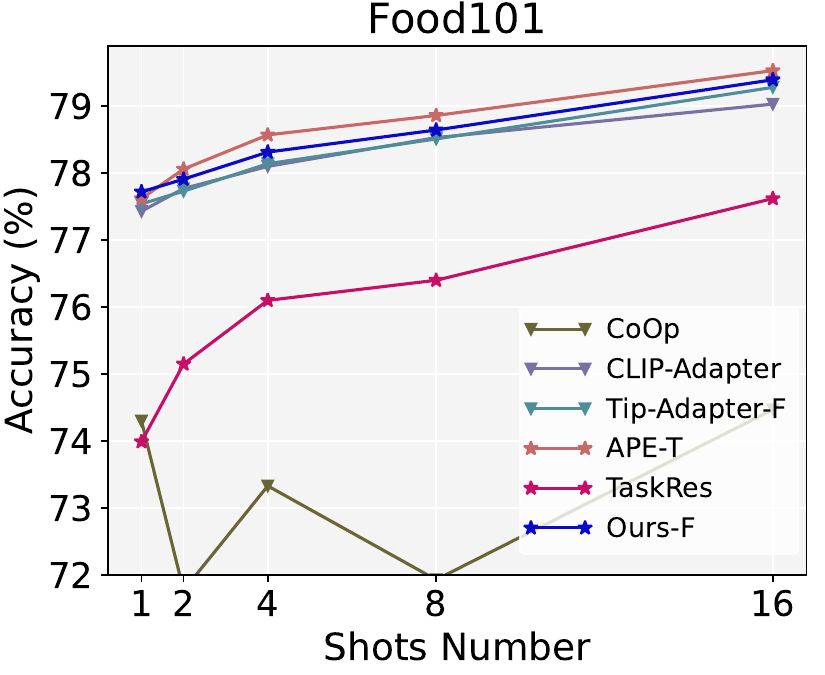}
	\end{minipage}
	\begin{minipage}{0.24\textwidth}
		\centering
		\includegraphics[width=1.0\textwidth]{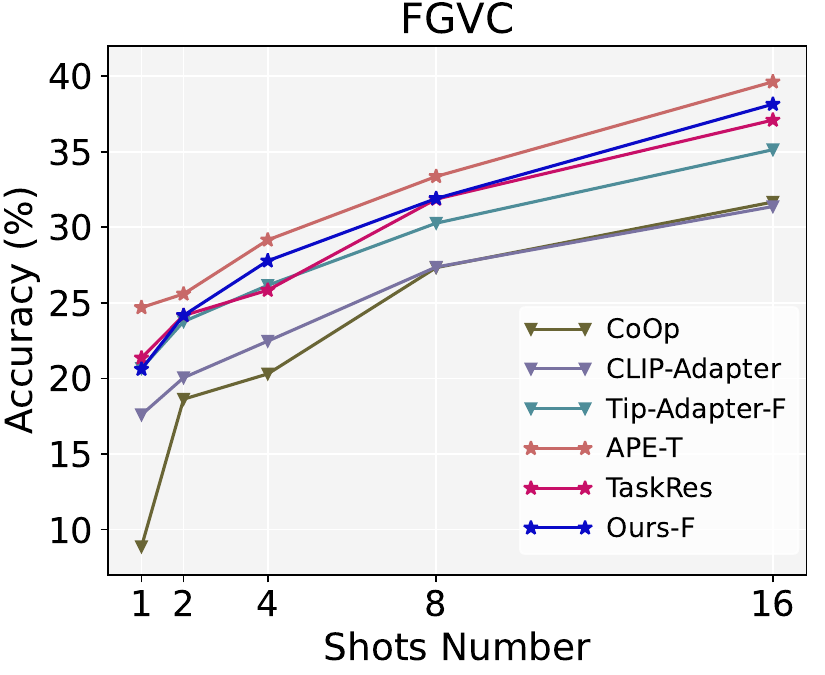}
	\end{minipage}
\\
	\begin{minipage}{0.24\textwidth}
		\centering
		\includegraphics[width=1.0\textwidth]{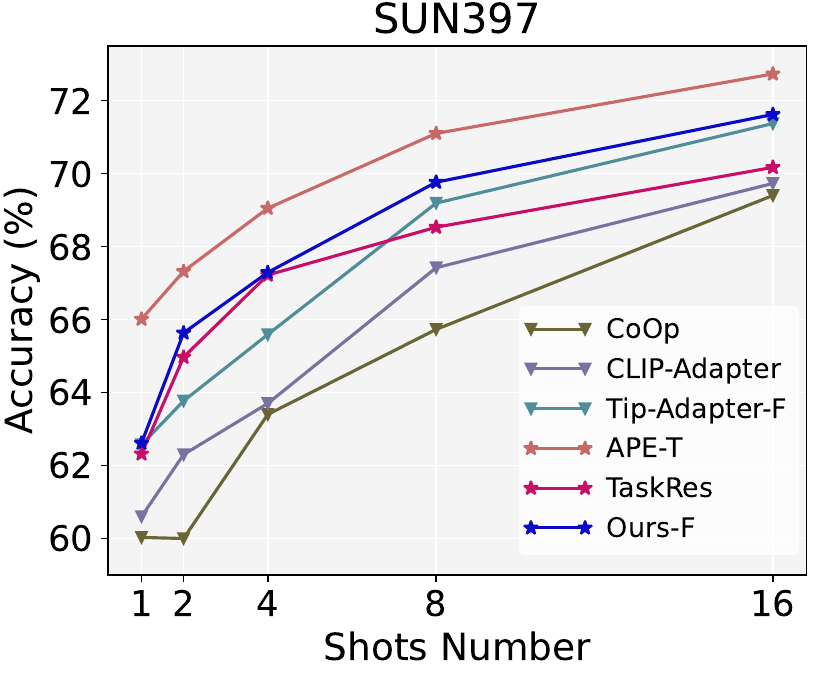}
	\end{minipage}
	\begin{minipage}{0.24\textwidth}
		\centering
		\includegraphics[width=1.0\textwidth]{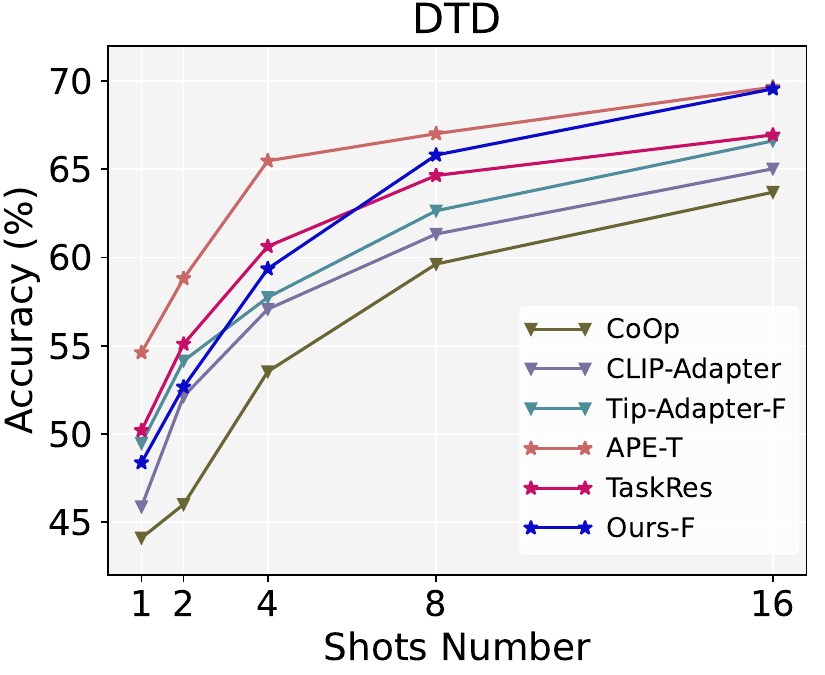}
	\end{minipage}
	\begin{minipage}{0.24\textwidth}
		\centering
		\includegraphics[width=1.0\textwidth]{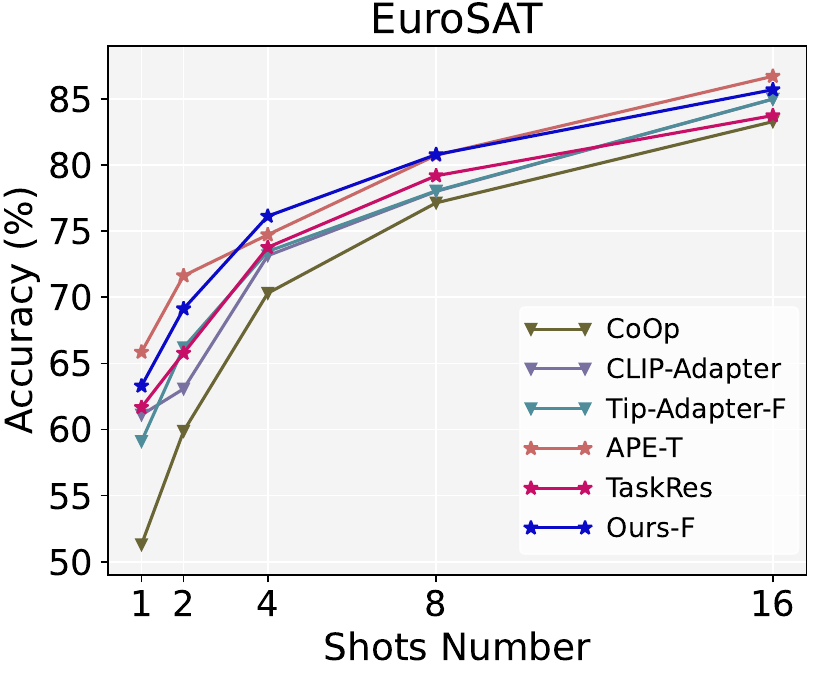}
	\end{minipage}
	\begin{minipage}{0.24\textwidth}
		\centering
		\includegraphics[width=1.0\textwidth]{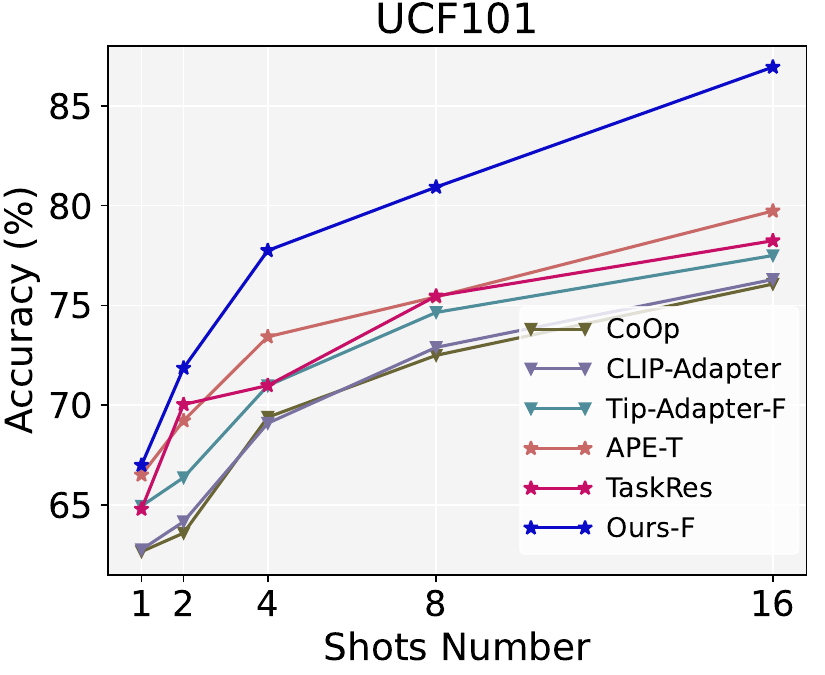}
	\end{minipage}
	\caption{
		{Comparison of training-required methods under the few-shot classification setting on the 11 datasets.}
	}
	\vspace{-10pt}
	\label{fig:train_fewshots}
\end{figure*}

\noindent\textbf{Training Details.} Following previous works, we adopt ResNet-50 as the default image encoder of CLIP. The effect of using different visual backbones will be studied later. In the similarity calibration stage, we use the linear projection layer, while the effect of using other structures will be studied. We randomly sample $p=500,000$ images from CC3M~\citep{CC3M} as the training set for similarity calibration. We use the SGD optimizer with weight decay of 5e-2 and momentum of 0.9 to optimize the parameters in calibration layers. The learning rate is 0.01 and the CosineAnnealingLR scheduler is utilized. The training batch size $b$ is 256 and the neighbor count $r$ is set to be 128. Besides, the training epoch is 10.

In adapting stage, following previous works, we search the best hyper-parameters on validation set. The searched parameters include $\alpha$, $\beta$, $\sigma$ and $\eta$. Besides, for the training-required mode, the training batch size is 256 and learning rate is 0.001. To reduce the complexity, we adopt the proposed group-based learning strategy on ImageNet and SUN397 for the 16-shot setting. For the other cases, we use the original extract computation. Lastly, all the experiments are run on a single NVIDIA GeForce RTX 3090.

\noindent\textbf{Evaluation Protocol.} For few-shot experiment, we adopt the identical protocol to previous works~\citep{TipAdapter}. For each dataset, we randomly select 1, 2, 4, 8, 16 training images per class to construct the training set. For training-free GPCache, we directly evaluate the performance on all test images of the same dataset. For training-required GPCache-F, we optimize the parameters on training set and subsequently evaluate the performance on test set. As for domain generalization setting, we train the model on the few-shot training set of ImageNet and evaluate the performance on ImageNetV2's or ImageNet-Sketch's test images.

\begin{table*}[!thb]
	\tabstyle{7pt}
	\caption{Few-shot classification results on 11 datasets with 16 shots.}
	\label{tab:few_shot}
	\renewcommand{\arraystretch}{1.1}
	\begin{tabular}{l |c c c c c c c c c c c |c}
		\toprule
		&  \rotbox{ImageNet} & \rotbox{Caltech101} & \rotbox{OxfordPets} & \rotbox{StanfordCars} & \rotbox{Flowers102} & \rotbox{Food101} & \rotbox{FGVCAircraft} & \rotbox{SUN397} & \rotbox{DTD} & \rotbox{EuroSAT} & \rotbox{UCF101} & \rotbox{Average}\\
		\midrule
		\multicolumn{4}{l}{\!\!\textBF{Training-free Methods:}}\\
		ZS-CLIP & 60.31 & 83.94 & 85.83 & 55.71 & 66.10 & 77.32 & 17.10 & 58.53 & 40.07 & 37.54 & 61.33 & 58.53\\ 
		CALIP & 60.57 & 87.71 & 86.21 & 56.27 & 66.38 & 77.42 & 17.76 & 58.70 & 42.39 & 38.90 & 61.72 & 59.46\\ 
		Tip-Adapter & 62.03 & 90.18 & 88.53 & 74.10 & 89.81 & 77.87 & 29.94 & 66.83 & 60.70 & 70.63 & 70.68 & 71.03\\ 
		Tip-X & 62.16 & 90.70 & \textBF{89.86} & 67.30 & 90.29 & 77.93 & 30.12 & 68.00 & 63.53 & 73.16 & 71.95 & 71.37\\ 
		APE & 63.42 & 92.49 & 88.88 & 70.31 & 91.96 & 78.50 & 31.26 & 69.72 & 67.73 & 78.16 & 74.49 & 73.36 \\ 
		GDA & 63.82 & 92.55 & 88.81 & 75.12 & 95.72 & 79.05 & \textBF{40.61}& \textBF{70.70} & 66.51 & \textBF{86.12} & 77.53 & 76.05\\ 
		\rowcolor{tabhighlight}
		GPCache & \textBF{64.17} & \textBF{93.56} & 89.82 & \textBF{76.96} & \textBF{96.64} & \textBF{79.45} & 40.56 & 70.22 & \textBF{69.58} & 84.63 & \textBF{78.61} & \textBF{76.75} \\
		\midrule
		\multicolumn{4}{l}{\!\!\textBF{Training-required Methods:}}\\
		CoOp & 62.97 & 92.07 & 86.30 & 73.33 & 94.73 & 74.47 & 31.67 & 69.40 & 63.70 & 83.27 & 76.07 & 73.45 \\ 
		TaskRes & 65.58 & 93.22 & 87.80 & 76.68 & 95.72 & 77.51 & 35.61 & 70.87 & 67.27 & 84.09& 77.43 & 75.62\\ 
		CLIP-Adapter & 63.70 & 92.39 & 88.73 & 74.10 & 94.10 & 79.03 & 31.37 & 69.73 & 65.03 & 84.97 & 76.30 & 74.50\\ 
		Tip-Adapter-F & 65.51 & 92.86 & 89.64 & 75.31 & 94.44 & 79.28 & 35.13 & 71.37 & 66.61 & 84.98 & 77.50 & 75.69\\ 
		APE-T & \textBF{66.07} & 93.14 & 90.24 & 76.53 & 96.14 & \textBF{79.53} & \textBF{39.62} & \textBF{72.73} & \textBF{69.66} & \textBF{86.70} & 79.73 & 77.28\\
		\rowcolor{tabhighlight}
		GPCache-F & 65.52 & \textBF{94.12} & \textBF{92.39} & \textBF{77.94} & \textBF{96.94} & 79.39 & 38.14 & 71.62 & 69.56 & 85.68 & \textBF{86.94} & \textBF{78.02}\\
		\bottomrule
	\end{tabular}
\end{table*}

\noindent\textbf{Compared Methods.} The proposed adaptation method does not learn prompt text, making it more efficient. As such, we mainly compare it with recent efficient tuning methods. In training-free case, we consider \textbf{six} methods: \textbf{zero-shot CLIP (ZS-CLIP)}~\citep{CLIP} is a method that constructs classifier with the prompt template ``A photo of a \{CLASS\}", \textbf{CALIP}~\citep{CALIP} is a method that incorporates parameter-free attention module for enhancing the zero-shot classification performance, \textbf{Tip-Adapter}~\citep{TipAdapter} is a method that combines the logits from zero-shot classifier and plain cache model, \textbf{Tip-X}~\citep{TipX} is a method that constructs cache model by retrieving images from LAION-5B~\citep{LAION5B} or generating images with Stable Diffusion model~\citep{LDM}, \textbf{APE}~\citep{APE} is a method introducing feature refinement to cache model, and \textbf{GDA}~\citep{GDA} is a method that exploits Gaussian Discriminant Analysis instead of cache model. In training-required case, we mainly consider \textbf{five} methods: \textbf{CoOp}~\citep{CoOp} is a method introducing learnabled prompt prefix to text inputs and optimizing them by back-propagation,  \textbf{CLIP-Adapter}~\citep{CLIPAdapter} is a method that introduces feature adapters on either visual or language branch, \textbf{Tip-Adapter-F}~\citep{TipAdapter} is a training-required variant of Tip-Adapter, \textbf{TaskRes}~\citep{TaskRes} is a method that freezes the zero-shot classifier and learns the task residual, and \textbf{APE-T}~\citep{APE} is a training-required variant of APE.

\subsection{Results on Few-shot Classification} The results with varying shots are shown in Fig.~\ref{fig:trainfree_fewshots} and Fig.~\ref{fig:train_fewshots}, corresponding to training-free and training-required cases, respectively. In terms of mean performance, GPCache achieves better accuracies compared to all other training-free methods and GPCache-F achieves comparable performance to APE-T in the training-required case. GPCache-F surpasses APE-T with 8 and 16 shots, but is outperformed by the latter under 1, 2 and 4 shots. Nonetheless, the training-free variant APE does not perform as well as our training-free variant GPCache. Besides, although GDA performs slightly worse than GPCache, how to extend GDA to the training-required case is still unclear.

Table~\ref{tab:few_shot} has shown the numerical results of different methods with 16 shots, under both training-free and training-required settings. For the training-free case, on average, GPCache outperforms the second best method GDA by 0.70\%, and outperforms the third best method APE by 3.39\%. Specifically, GPCache achieves better accuracy than GDA on 8 out of 11 datasets, and achieves better accuracy than APE all datasets. As for training-required case, on average, GPCache-F outperforms the second best method APE-T by 0.74\%.

\subsection{Results on Domain Generalization}
The results of domain generalization are listed in Table~\ref{tab:dg}. For the training-free case, GPCache significantly outperforms Tip-Adapter on both source domain (+1.70\% on ImageNet) and target domains (+1.88\% on ImageNet-V2 and +0.38\% on ImageNet-Sketch). As for training-required case, GPCache-F outperforms Tip-Adapter-F on ImageNet and ImageNet-Sketch by 0.19\% and 0.50\%, respectively, but achieves comparable accuracy on ImageNet-V2.

\begin{table}[!thb]
	\tabstyle{2pt}
	\caption{Results of domain generalization (16 shots).}
	\label{tab:dg}
	\renewcommand{\arraystretch}{1.2}
	\begin{tabular}{l c cc}
		\toprule
		& Source & \multicolumn{2}{c}{Target} \\
		\cmidrule(lr){2-2} \cmidrule(lr){3-4}
		& {ImageNet} & {ImageNet-V2} & {ImageNet-Sketch} \\
		\midrule
		\multicolumn{4}{l}{\!\!\textBF{Training-free Methods:}}\\
		ZS-CLIP & 60.33 & 53.27 & 35.44\\
		CALIP & 60.57 & 53.70 & 35.61 \\
		Tip-Adapter & 62.47 & 54.46 & 35.85 \\
		\rowcolor{tabhighlight}
		GPCache & \textBF{64.17} & \textBF{56.34} & \textBF{36.23} \\
		\midrule
		\multicolumn{4}{l}{\!\!\textBF{Training-required Methods:}}\\
		CoOp & 62.95 & 54.58 & 31.04 \\
		CLIP-Adapter & 63.59 & 55.69 & 35.68 \\
		Tip-Adapter-F & 65.33 & \textBF{56.49} & 35.96 \\
		\rowcolor{tabhighlight}
		GPCache-F & \textBF{65.52} & 56.36 & \textBF{36.46} \\
		\bottomrule
	\end{tabular}
\end{table}

\subsection{Ablation Study}
\textbf{Effectiveness of Calibration.} To validate the effectiveness of three calibration modules, we use the Tip-Adapter as the baseline, and add different modules gradually. The results with different configurations are shown in Table~\ref{tab:calibration}. By comparing the results of B+W and B, we can conclude that the proposed weight calibration contributes to improved performance. Upon comparing the outcomes of B+W+C and B+W, notable accuracy enhancements of 1.80\% and 0.63\% are observed in the training-free and training-required scenarios, respectively. These improvements underscore the efficacy of incorporating confidence calibration. By comparing the results of B+S and B, we can conclude that the similarity calibration is needed. While adding similarity calibration to the setting of B+W+C, improved accuracies can also be observed, further validating the effectiveness of similarity calibration.

\begin{table}[!t]
	\tabstyle{2pt}
	\caption{\textbf{Effectiveness of three calibration modules.} The second the third columns are the average results on 11 datasets. \textbf{B}: the Tip-Adapter(-F) baseline, \textbf{S}: similarity calibration, \textbf{W}: weight calibration, \textbf{C}: confidence calibration. }
	\label{tab:calibration}
	\renewcommand{\arraystretch}{1.15}
	\begin{tabular}{l cc}
		\toprule
		& training-free & training-required \\
		\midrule
		\textbf{Baseline:}&&\\
		B & 70.35 & 75.53 \\
		\midrule
		\textbf{Weight Calibration:}&&\\
		B$+$W & 74.38 & 76.58\\
		{B$+$W $-$ B}&\textBF{+4.03} & \textBF{+1.05} \\
		\midrule
		\textbf{Confidence Calibration:}&&\\
		B$+$W$+$C & 76.18 & 77.21 \\
		{B$+$W$+$C $-$ B$+$W}&\textBF{+1.80} & \textBF{+0.63} \\
		\midrule
		\textbf{Similarity Calibration:}&&\\
		B$+$S & 71.12 & 76.01 \\
		{B$+$S $-$ B}&\textBF{+0.77} & \textBF{+0.48} \\
		B$+$W$+$C$+$S& 76.75 & 78.02 \\
		{B$+$W$+$C$+$S $-$ B$+$W$+$C}&\textBF{+0.57} & \textBF{+0.81} \\
		\bottomrule
	\end{tabular}
\vspace{-4pt}
\end{table}
\begin{table}[!tb]
	\tabstyle{9pt}
	\caption{\textbf{Effectiveness of hard mining.} The second row denotes the method without hard mining. $p$ is set to be 10,000.}
	\label{tab:hard_mine}
	\renewcommand{\arraystretch}{1.15}
	\begin{tabular}{l cc}
		\toprule
		$r$ & training-free & training-required \\
		\midrule
		-- & 76.33 & 77.63 \\
		16 & 76.34 & 77.72 \\
		32 & 76.38 & 77.52 \\
		64 & \textBF{76.48} & 77.72 \\
		128 & 76.40 & \textBF{77.79} \\
		256 & 76.17 & 77.50 \\
		\bottomrule
	\end{tabular}
\vspace{-5pt}
\end{table}

\noindent\textbf{Hard Mining.} The results with and without hard mining are shown in Table~\ref{tab:hard_mine}. From the table, we can see that the hard samples allow us to obtain a better image-image similarity function, reflected by the improved accuracy in both the training-free and training-required cases. We believe that the hard samples can help the model capture more discriminative similarity. Additionally, we discern a trend where the downstream performance tends to decline as the number of $r$ increases. This can be attributed to the introduction of more simple samples when $r$ is larger, thereby diluting the positive influence of hard samples on the overall performance.

\begin{table}[!tb]
	\tabstyle{1.4pt}
	\caption{\textbf{Comparison of the structure of calibration layer.} In $Linear_{q}()$, $q$ denotes reducing the input dimension to $1/q$. In this experiment, $p$ is set to be 10,000.}
	\label{tab:calib_layer}
	\renewcommand{\arraystretch}{1.15}
	\begin{tabular}{l ccc}
		\toprule
		Proj & residual & training-free & training-required \\
		\midrule
		$Linear()$ &$\checkmark$& \textBF{76.40} & \textBF{77.79} \\
		$Linear()$ & $\times$& 76.31 & 77.05 \\
		$Linear_{\frac{1}{2}}(Linear_2())$ &$\checkmark$& 76.23 & 77.57 \\
		$Linear_{\frac{1}{4}}(Linear_4())$ &$\checkmark$& 76.22 & 77.64 \\
		$Linear_{\frac{1}{8}}(Linear_8())$ &$\checkmark$& 76.04 & 77.45 \\
		\bottomrule
	\end{tabular}
\end{table}
\begin{figure}[!tb]
	\centering
	\begin{subfigure}{.48\linewidth}
		\centering
		\includegraphics[width=\linewidth]{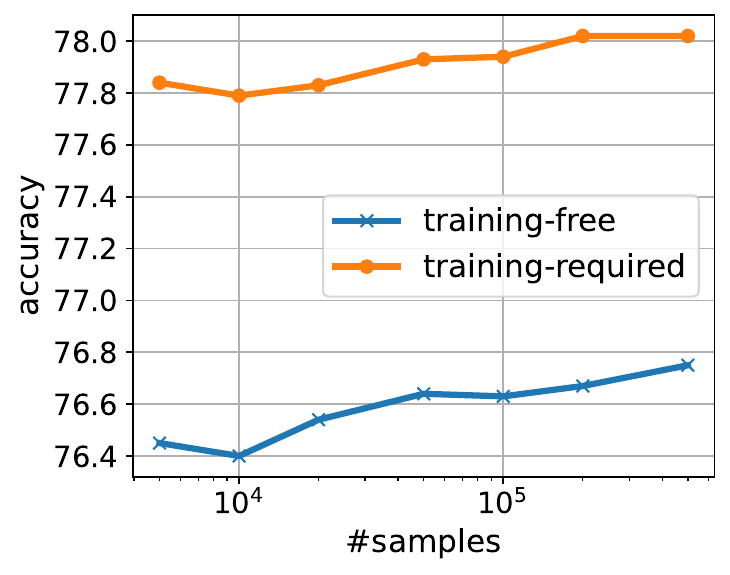}
		\caption{}
		\label{fig:effect_K}
	\end{subfigure}%
	\hspace{2pt}
	\begin{subfigure}{.5\linewidth}
		\centering
		\includegraphics[width=\linewidth]{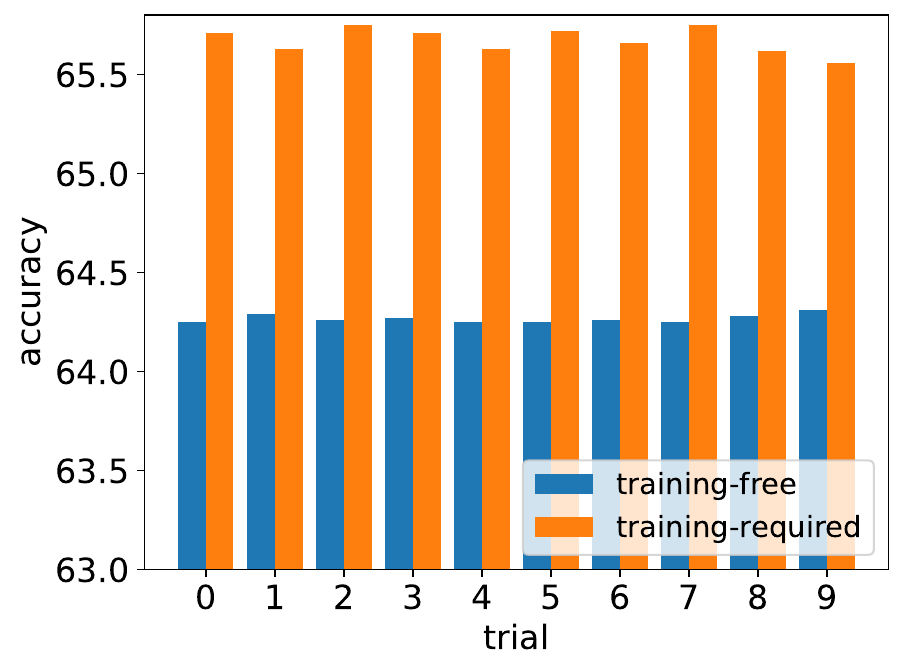}
		\caption{}
		\label{fig:robust_group_split}
	\end{subfigure}
	\caption{(a) Effect of number of unlabeled training samples. (b) Robustness to different group splitting in approximating GP.}
	\vspace{-5pt}
\end{figure}

\noindent\textbf{Calibration Layers.} The structure of calibration layers is ablated in Table~\ref{tab:calib_layer}. As can be seen, it is important to add the residual connection. We conjecture that the residual connection can maintain the pre-trained knowledge. We have also tried a low-rank structure like that in LoRA~\citep{LoRA}, however, we cannot observe any improvement. 

\noindent\textbf{Number of Unlabeled Training Samples.} The average accuracy curve over 11 datasets with increasing number of unlabeled training samples is shown in Fig.~\ref{fig:effect_K}. As can be seen from the figure, with more unlabeled training samples, we can obtain a better calibrated similarity function, which will help to strengthen the performance of cache model.

\begin{table}[!tb]
	\tabstyle{3pt}
	\caption{Comparison of training time with different approximation methods on ImageNet. $c$ denotes the number of classes. For Nystr{\"{o}}m Approximation, we sample the same number of training samples per class. We find that, Nystr{\"{o}}m Approximation with larger $L$ tends to fail to converge while solving the SVD problem.}
	\label{tab:complexity}
	\renewcommand{\arraystretch}{1.1}
	\begin{tabular}{l cccc}
		\toprule
		& \multicolumn{2}{c}{training-free} & \multicolumn{2}{c}{training-required} \\
		& accuracy & time/s & accuracy & time/s \\
		\midrule
		Mean & 63.72 & 37 & 64.23 & 1345 \\
		\midrule
		Nystr{\"{o}}m ($L=c$) & 62.18 & 133 & 65.55 & 1565 \\
		Nystr{\"{o}}m ($L=2c$) & 62.18 & 258 & -- & -- \\
		Nystr{\"{o}}m ($L=4c$) & 62.23 & 904 & -- & --\\
		\midrule
		RFF ($D=256$) & 62.08 & 75 & 61.91 & 491 \\
		RFF ($D=512$) & 62.21 & 85 & 62.87 & 432 \\
		RFF ($D=1024$) & 62.34 & 108 & 63.63 & 477 \\
		RFF ($D=2048$) & 62.46 & 160 & 64.46 & 874 \\
		RFF ($D=4096$) & 62.51 & 298 & 65.09 & 3649 \\
		\midrule
		ours ($g=8$) & 64.28 & 114 & 65.74 & 4829 \\
		ours ($g=16$) & 64.08 & 87 & 65.55 & 2902 \\
		ours ($g=32$) & 63.87 & 76 & 64.59 & 2218 \\
		ours ($g=64$) & 63.53 & 74 & 64.34 & 2092 \\
		\midrule
		ours ($g=8$ nograd) & 64.28 & 114 & 64.97 & 1513 \\
		ours ($g=16$ nograd) & 64.08 & 87 & 65.00 & 1447 \\
		ours ($g=32$ nograd) & 63.87 & 76 & 65.23 & 1464 \\
		ours ($g=64$ nograd) & 63.53 & 74 & 65.04 & 1533 \\
		\bottomrule
	\end{tabular}
\end{table}
\begin{table}[!tb]
	\tabstyle{6pt}
	\caption{Results with different backbones (16 shots).}
	\label{tab:backbone}
	\renewcommand{\arraystretch}{1.1}
	\begin{tabular}{l cccc}
		\toprule
		Model& ResNet-101& ViT-B/32 & ViT-B/16 \\
		\midrule
		\multicolumn{2}{l}{\textBF{Training-free Methods:}}\\
		ZS-CLIP & 59.86 & 61.88 & 65.23 \\
		Tip-Adapter & 71.96 & 72.37 & 76.80\\
		APE & 74.69& 75.19 & 79.34 \\
		GDA & 78.06 & 78.06 & 81.85 \\
		\rowcolor{tabhighlight}
		GPCache & \textBF{78.58}& \textBF{78.49}& \textBF{82.01}\\
		\midrule
		\multicolumn{2}{l}{\textBF{Training-required Methods:}}\\
		CoOp & 75.96 & 75.70 & 79.71 \\
		Tip-Adapter-F & 78.53 & 78.42 & 81.79\\
		APE-T & 78.84& 78.72 & 82.21 \\
		\rowcolor{tabhighlight}
		GPCache-F & \textBF{79.55}&\textBF{79.58} & \textBF{82.69}\\
		\bottomrule
	\end{tabular}
\end{table}

\noindent\textbf{Robustness to Group Splitting.} We validate the robustness of group splitting in group-based learning strategy for speeding up GPs in Fig.~\ref{fig:robust_group_split} . The results on ImageNet with 10 different seeds for splitting the groups are reported. The accuracy variances of both training-free and training-required modes are quite small, i.e., 0.02\% and 0.06\%, respectively, implying the robustness to group splitting.

\noindent\textbf{GP Approximation.} We compare the computational efficiency of different methods for approximating GPR in Table~\ref{tab:complexity}. The compared methods include: 1) \textbf{Mean}: averaging the image features of each class and then applying the exact GPR; 2) \textbf{Nystr{\"{o}}m}: applying Nystr{\"{o}}m Approximation with sampled training subset of size $L$; 3) \textbf{RFF}: applying the Random Fourier Features with transformation dimension $D$; 4) \textbf{ours}: the proposed method with the number of groups $g$. Besides, a variant of our method by turning off the gradients of the precision matrix is also compared. For Nystr{\"{o}}m Approximation and RFF, we use the following property of inverting a matrix for speedup: $(\mathbb{I}+ZZ^T)^{-1}=\mathbb{I}-Z(\mathbf{I}+Z^TZ)^{-1}Z^T$, where $Z\in\mathbb{R}^{d\times r}$ with $d>r$. From Table~\ref{tab:complexity}, we have the following observations:

\begin{figure}
	\centering
	\begin{subfigure}{.47\linewidth}
		\centering
		\includegraphics[width=\linewidth]{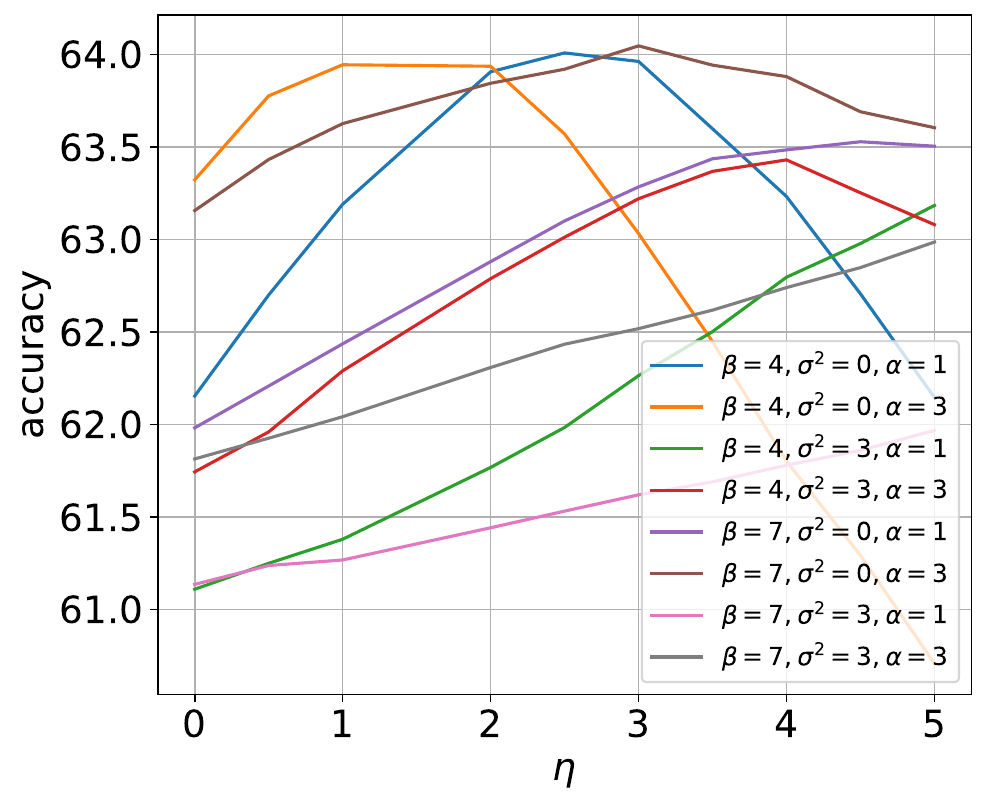}
		\caption{}
		\label{fig:effect_eta_sigma_sfig1}
	\end{subfigure}%
	\hspace{2pt}
	\begin{subfigure}{.47\linewidth}
		\centering
		\includegraphics[width=\linewidth]{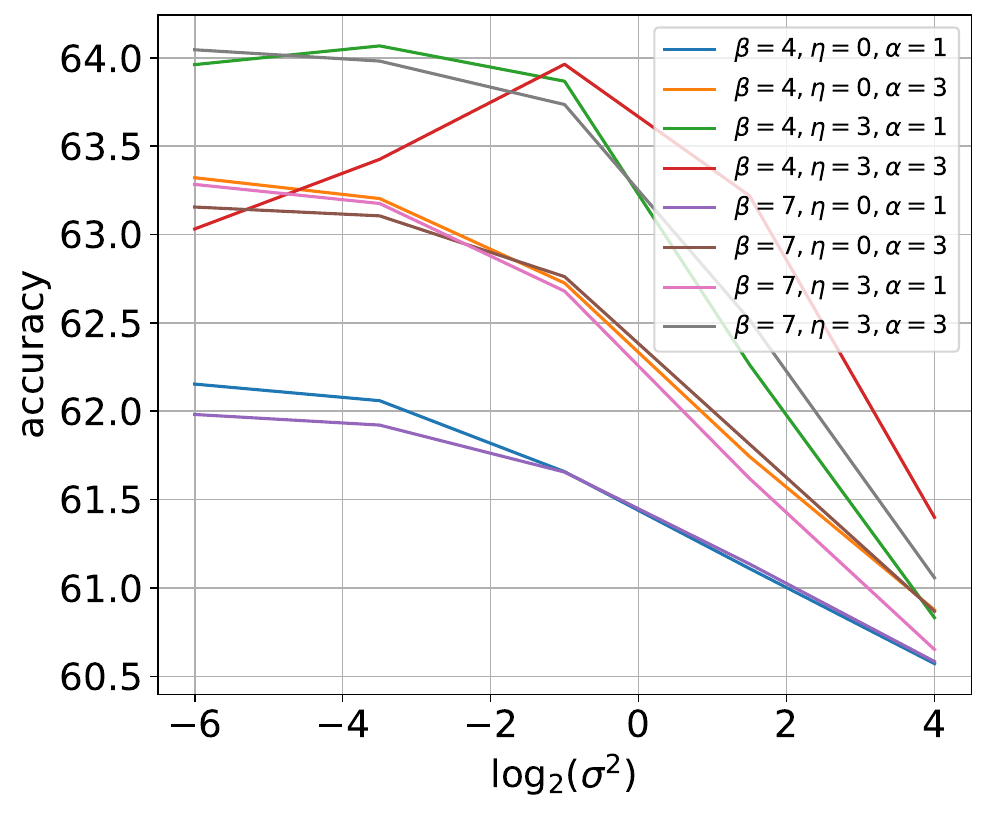}
		\caption{}
		\label{fig:effect_eta_sigma_sfig2}
	\end{subfigure}
	\\
	\begin{subfigure}{.47\linewidth}
		\centering
		\includegraphics[width=\linewidth]{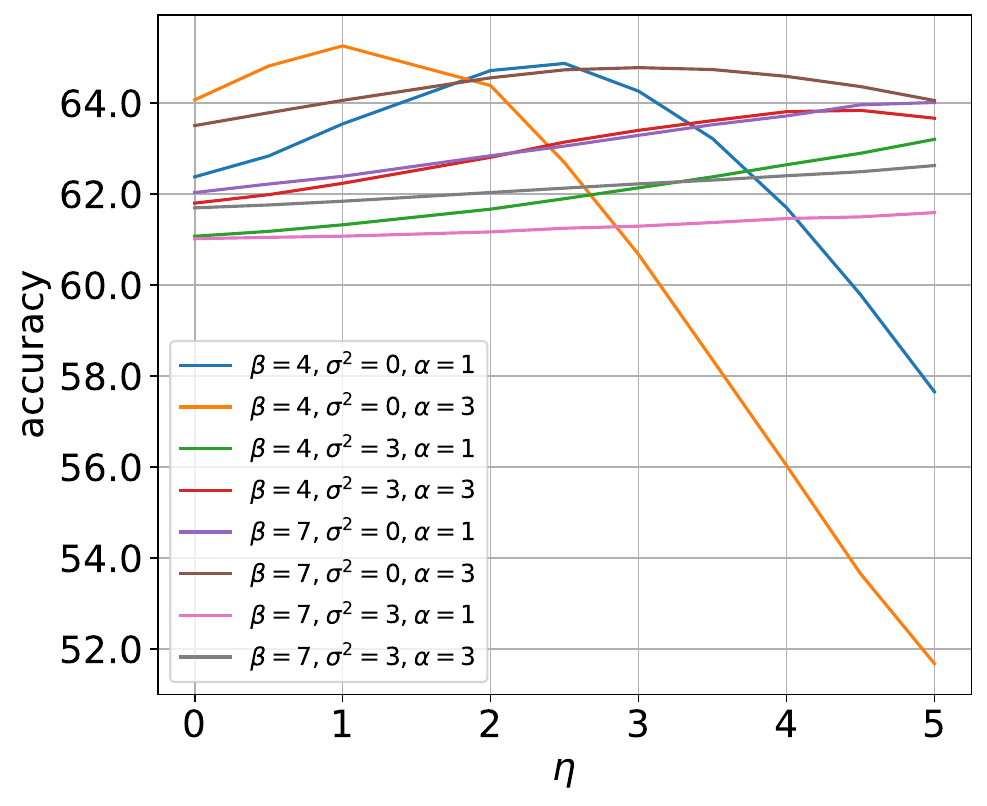}
		\caption{}
		\label{fig:effect_eta_sigma_sfig3}
	\end{subfigure}%
	\hspace{2pt}
	\begin{subfigure}{.47\linewidth}
		\centering
		\includegraphics[width=\linewidth]{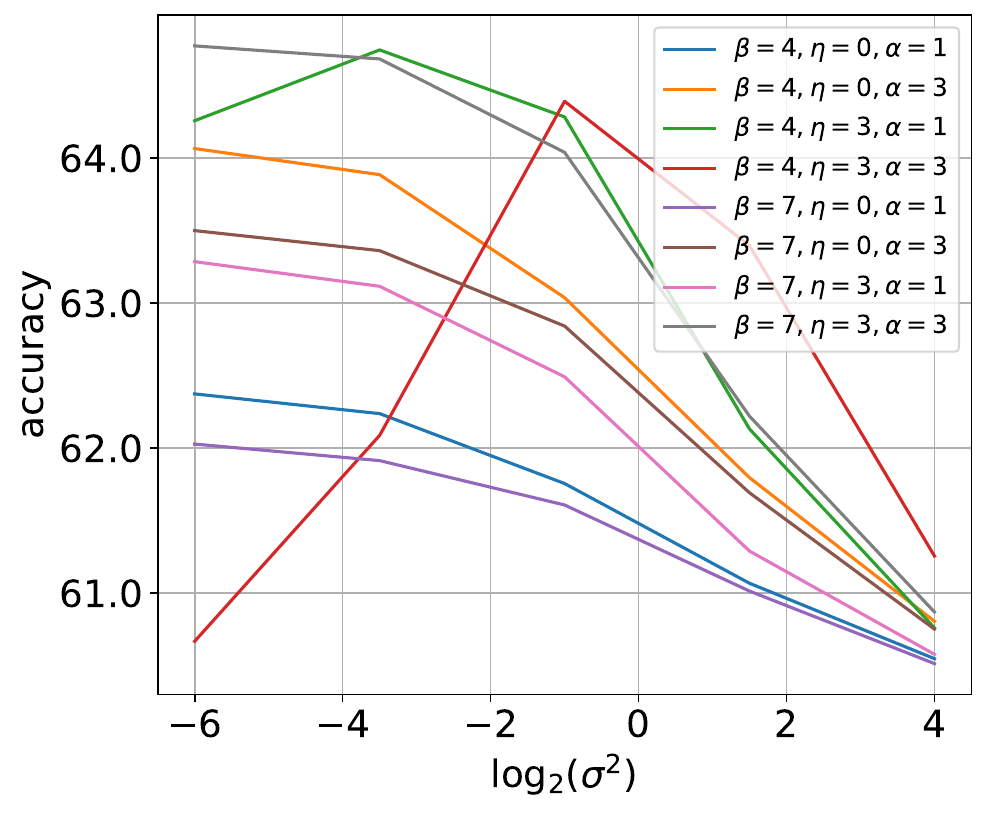}
		\caption{}
		\label{fig:effect_eta_sigma_sfig4}
	\end{subfigure}
	\caption{Effect of $\eta$ and $\sigma^2$ on validation set of ImageNet. Firs row: GPCache, second row: GPCache-F.}
	\label{fig:effect_eta_sigma}
\end{figure}

First, the Mean method serves a good baseline as it outperforms Nystr{\"{o}}m and RFF in the training-free case and outperforms RFF with small $D$s in the training-required case. Second, the Nystr{\"{o}}m Approximation works well in the training-required case, but does not performs as well in the training-free case. Third, only when higher transformation dimension is used can RFF acquire good results. The underlying reason is that RFF is a data-independent approximation method. Fourth, the proposed method can acquire the best accuracy with the shortest time in the training-free case, it also achieves comparable accuracy to Nystr{\"{o}}m with close training time. Lastly, while turning off the gradients of the precision matrix, the training speed can be boosted significantly without sacrificing the accuracy too much. In summary, the proposed group-based method is an effective and efficient method for speeding up GPs for vision-language model adaptation.

\noindent\textbf{Different Backbones.} To demonstrate the generalization ability of the proposed method to different image features, we compare the results of different methods using different backbones in Table~\ref{tab:backbone}. From the table, we can see that GPCache is compatible to other backbones like ViT-B/16. The improvements of GPCache and GPCache-F over the second best methods are notable, i.e., +0.52\%, +0.43\%, +0.16\% for ResNet-101, ViT-B/32 and ViT-B/16 respectively in the training-free case, and +0.71\%, +0.86\%, +0.48\% for ResNet-101, ViT-B/32 and ViT-B/16 respectively in the training-required case.

\noindent\textbf{Ablation on $\eta$ and $\sigma^2$.} The parameter $\eta$ acts as tuning the distribution of the variance and the noise variance $\sigma^2$ reflects the prior of the observation's accuracy. We study their effect on classification performance on validation sets in Fig.~\ref{fig:effect_eta_sigma}. When $\eta$ approaching to zero, the effect of confidence based weighting will be weaken, leading to sub-optimal results. This implies that it is important to introduce confidence calibration. As $\sigma^2$ approaches to $+\infty$, the GPCache will become a plain cache model. The results in Fig.~\ref{fig:effect_eta_sigma_sfig2} and Fig.~\ref{fig:effect_eta_sigma_sfig4} reflect that the weight calibration through GPR is helpful.

\section{Conclusion}
This paper identifies the core problems of the existing cache model in prevalent vision-language adaptation methods, i.e., image-image similarity gap between pre-training and fine-tuning, less expressive weight function and the lack of uncertainty estimation. Accordingly, we propose similarity calibration, weight calibration and confidence calibration to address these issues. The similarity calibration follows the self-supervised learning paradigm and optimizes added calibration layers on unlabeled images using contrastive loss. The weight calibration is implemented by introducing the precision matrix and the confidence calibration is implemented by re-scaling the cache model's logits by the inverse of predictive variance. With these calibration modules, both training-free and training-required variants are proposed. The experiments under the few-shot classification and domain generalization settings validate the effectiveness of our methods. As an extra contribution, we proposed a group-based learning strategy, which makes the GPs be more tractable while being used in vision-language model adaptation. Our ablation experiments demonstrate the effectiveness. 

Our method is not without limitations. Although we have put in a lot effort in cutting down the complexity of GPs, the time complexity in the training-required case is still relatively high, particularly with small number of groups. In future, we will investigate more advanced methods in reducing complexity.

\section*{Data Availability Statements}
The ImageNet dataset is available at \href{https://image-net.org/index.php}{\nolinkurl{https://image-net.org/index.php}}. The Caltech101 dataset is available at \href{http://www.vision.caltech.edu/datasets}{\nolinkurl{http://www.vision.caltech.edu/datasets}}. The OxfordPets, Flowers102, DTD and FGVCAircraft datasets are available at \href{https://www.robots.ox.ac.uk/~vgg/data}{\nolinkurl{https://www.robots.ox.ac.uk/~vgg/data}}. The StanfordCars dataset is available at \href{http://ai.stanford.edu/~jkrause}{\nolinkurl{http://ai.stanford.edu/~jkrause}}. The Food101 dataset is available at \href{https://data.vision.ee.ethz.ch/cvl/datasets_extra/food-101}{ \nolinkurl{https://data.vision.ee.ethz.ch/cvl/datasets\_extra/food-101}}. The SUN397 dataset is available at \href{https://vision.princeton.edu/projects/2010/SUN}{\nolinkurl{https://vision.princeton.edu/projects/2010/SUN}}. The EuroSAT dataset is available at \href{http://madm.dfki.de/downloads}{\nolinkurl{http://madm.dfki.de/downloads}}. The UCF101 dataset is available at \href{https://www.crcv.ucf.edu/data/UCF101.php}{\nolinkurl{https://www.crcv.ucf.edu/data/UCF101.php}}. The ImageNet-V2 dataset is available at \href{https://github.com/modestyachts/ImageNetV2}{\nolinkurl{https://github.com/modestyachts/ImageNetV2}}. The ImageNet-Sketch dataset is available at \href{https://github.com/HaohanWang/ImageNet-Sketch}{\nolinkurl{https://github.com/HaohanWang/ImageNet-Sketch}}.

\begin{acknowledgements}
This research was supported by the National Natural Science Foundation of China under Grant 62306310.
\end{acknowledgements}

%
%

\bibliographystyle{spbasic}      
\bibliography{main}   

\end{document}